\documentclass[10pt,twocolumn,letterpaper]{article}

\usepackage{cvpr}              
 \usepackage{indentfirst}
 \usepackage{mathtools}
\usepackage{subcaption, caption}
\usepackage[sort&compress]{natbib}

\usepackage{booktabs}       
\usepackage{amsfonts}       
\usepackage{nicefrac}       
\usepackage{microtype}      

\usepackage{graphicx}
\usepackage{amsmath}
\usepackage{amssymb}

\usepackage{diagbox}
\usepackage{multicol}
\usepackage{enumerate}
\usepackage{times}
\usepackage{epsfig}
\usepackage{threeparttable}
\usepackage{enumitem}
\usepackage{multirow}
\usepackage{color}
\usepackage{array}
\usepackage{setspace}
\usepackage{makecell}

\usepackage[dvipsnames]{xcolor}

\definecolor{cvprblue}{rgb}{0.21,0.49,0.74}
\usepackage[pagebackref,breaklinks,colorlinks,citecolor=cvprblue]{hyperref}

\def\paperID{2209} 
\def\confName{CVPR}
\def\confYear{2024}

\title{Seeing Motion at Nighttime with an Event Camera}

\author{Haoyue Liu\textsuperscript{1}, Shihan Peng\textsuperscript{1}, Lin Zhu\textsuperscript{2}, Yi Chang\textsuperscript{1}\footnotemark[1], Hanyu Zhou\textsuperscript{1}, Luxin Yan\textsuperscript{1}\\
\textsuperscript{1} National Key Lab of Multispectral Information Intelligent Processing Technology\\
School of Artificial Intelligence and Automation, Huazhong University of Science and Technology, China\\
\textsuperscript{2}Beijing Institute of Technology\\
{\tt\small \{liuhy, pengshihan, yichang, hyzhou, yanluxin\}@hust.edu.cn, \tt\small linzhu@bit.edu.cn}}

\begin{document}

\twocolumn[{%
\renewcommand\twocolumn[1][]{#1}%
\maketitle
\begin{center}
    \centering

    \captionsetup{type=figure}
	\subcaptionbox{\scriptsize Conventional camera}{\includegraphics[width = 0.249\linewidth]{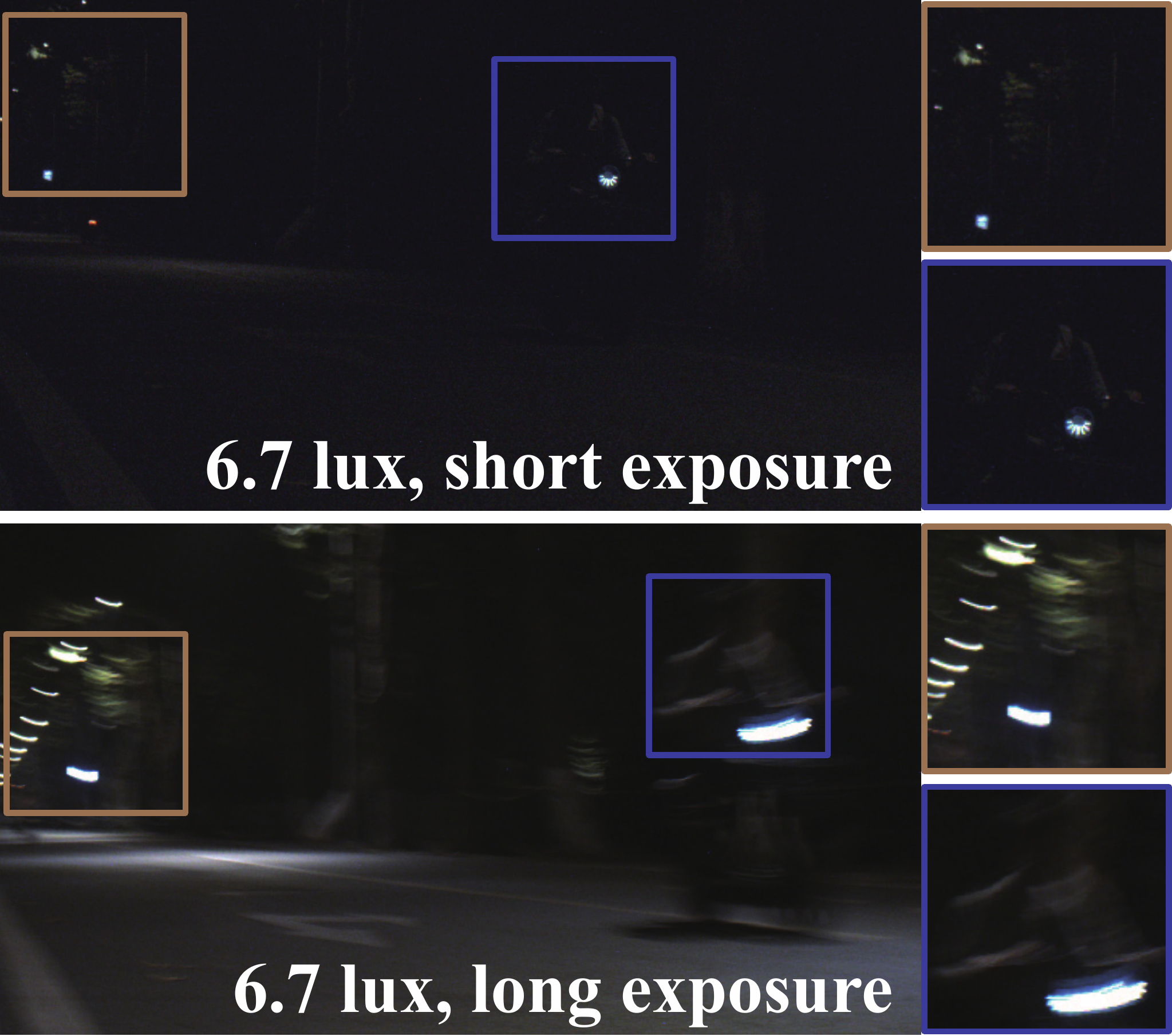}}\hfill
	\subcaptionbox{\scriptsize Event camera} {\includegraphics[width = 0.249\linewidth]{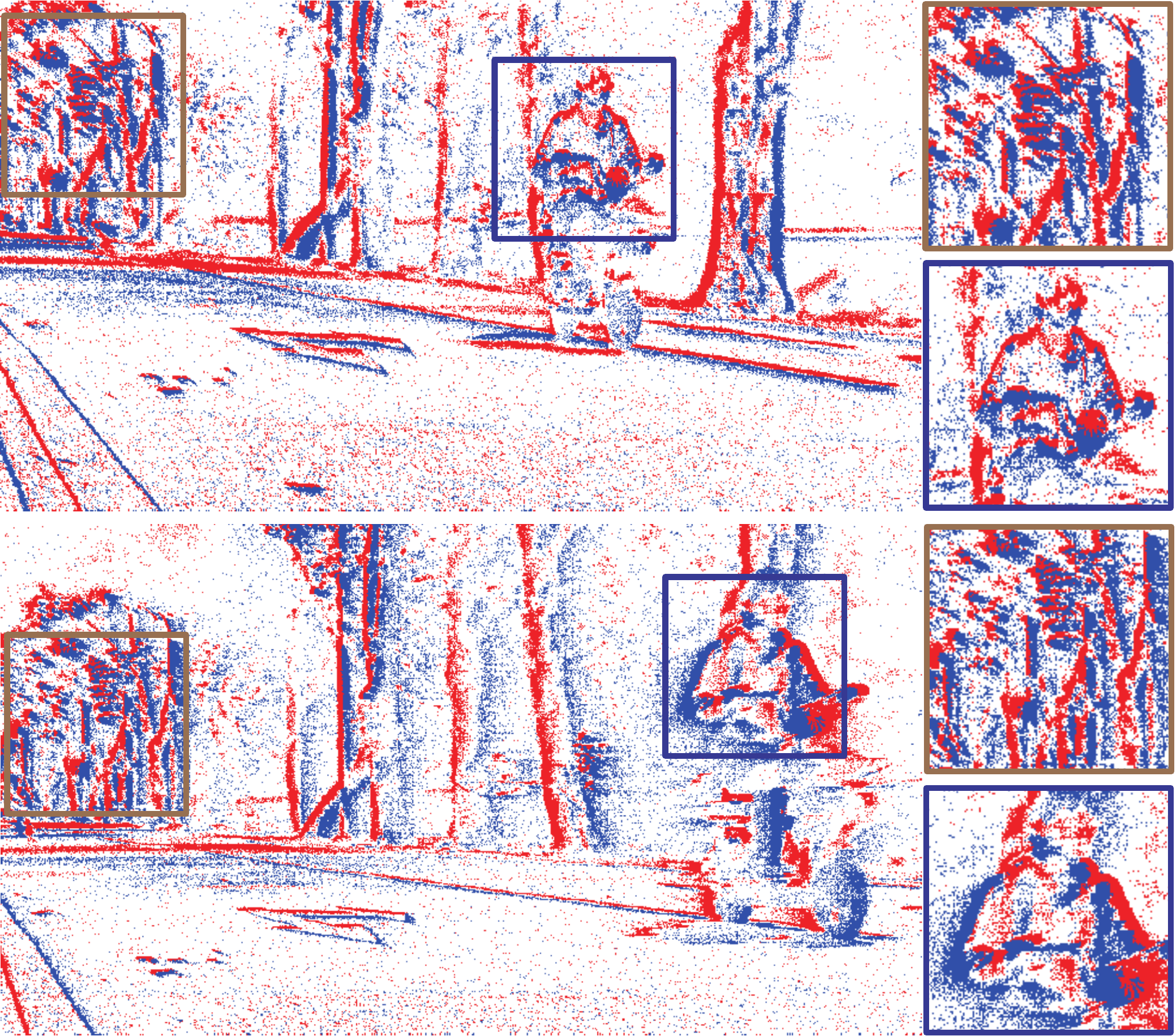}}\hfill
	\subcaptionbox{\scriptsize Reconstruction from E2VID+ \cite{stoffregen2020reducing}} {\includegraphics[width = 0.249\linewidth]{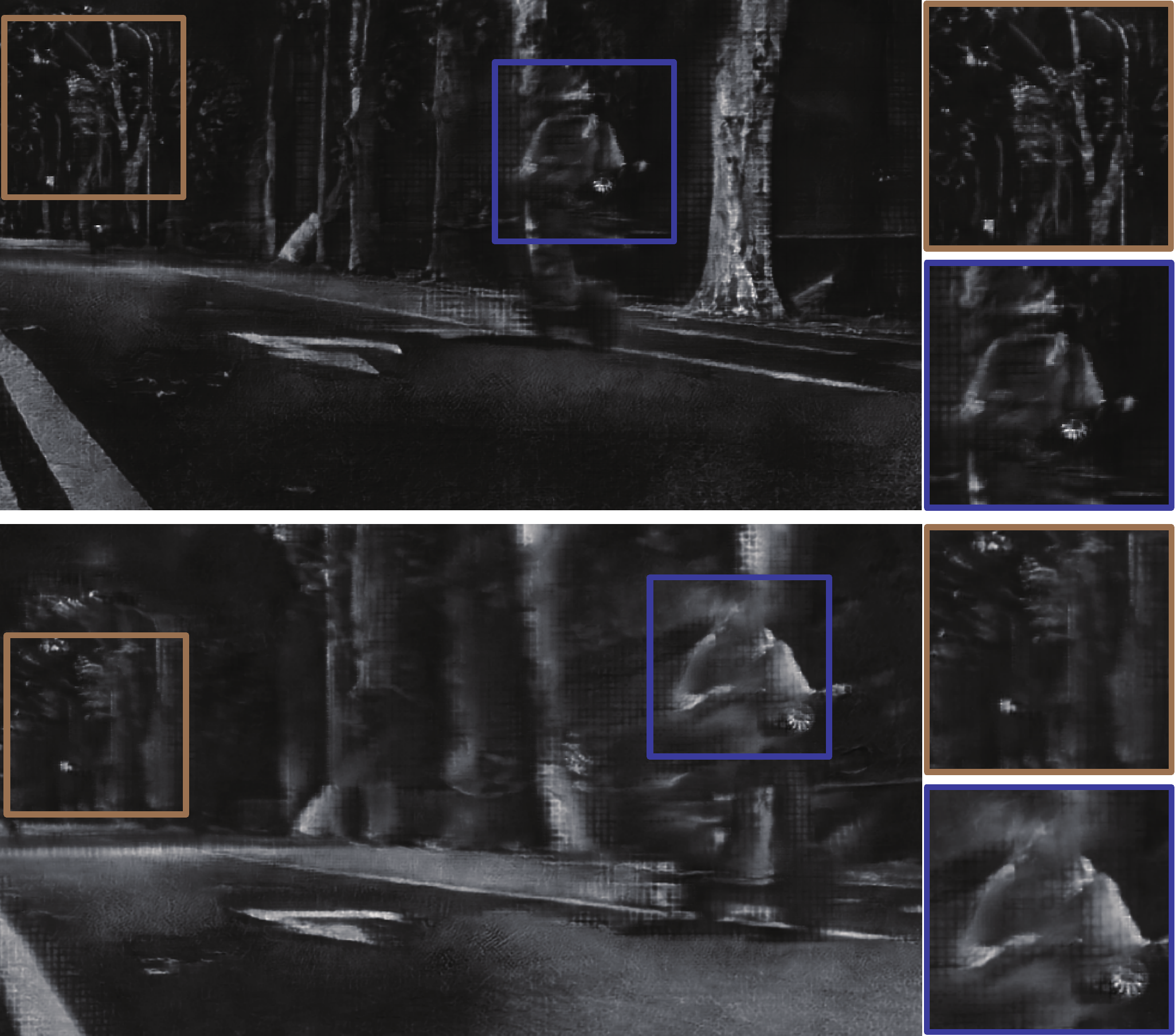}}\hfill
	\subcaptionbox{\scriptsize Reconstruction from NER-Net~(ours)}{\includegraphics[width =0.249\linewidth]{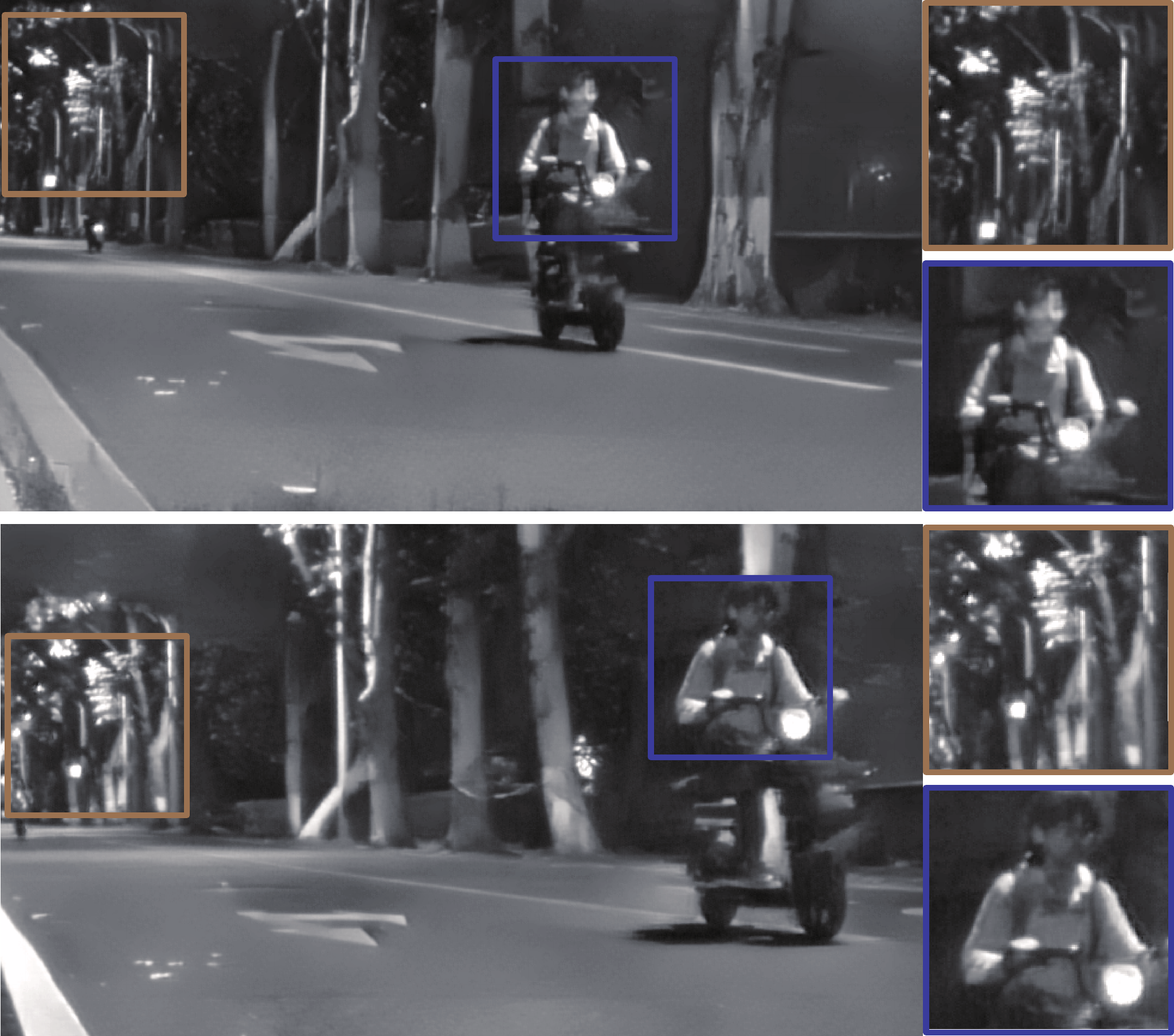}}
    \caption{Comparison between frame-based imaging and event-based imaging. (a) The extended exposure time of conventional cameras leads to motion blur in low-light conditions (\textbf{6.7 lux}). (b) Event cameras can capture high-speed and high-dynamic-range scene information. (c) and (d) are reconstruction results of E2VID+ \cite{stoffregen2020reducing} and the proposed method.}
 \label{fig1_imgaing_example}
\end{center}%
}]

\begin{abstract}
\vspace{-0.4cm}
We focus on a very challenging task: imaging at nighttime dynamic scenes. Most previous methods rely on the low-light enhancement of a conventional RGB camera. However, they would inevitably face a dilemma between the long exposure time of nighttime and the motion blur of dynamic scenes. Event cameras react to dynamic changes with higher temporal resolution (microsecond) and higher dynamic range (120dB), offering an alternative solution. In this work, we present a novel nighttime dynamic imaging method with an event camera. Specifically, we discover that the event at nighttime exhibits temporal trailing characteristics and spatial non-stationary distribution. Consequently, we propose a nighttime event reconstruction network (NER-Net) which mainly includes a learnable event timestamps calibration module (LETC) to align the temporal trailing events and a non-uniform illumination aware module (NIAM) to stabilize the spatiotemporal distribution of events. Moreover, we construct a paired real low-light event dataset (RLED) through a co-axial imaging system, including 64,200 spatially and temporally aligned image GTs and low-light events. Extensive experiments demonstrate that the proposed method outperforms state-of-the-art methods in terms of visual quality and generalization ability on real-world nighttime datasets. The project are available at: \url{https://github.com/Liu-haoyue/NER-Net}.
\end{abstract}
\vspace{-0.5cm}
\section{Introduction}
\label{sec:intro}
\indent
Imaging in nighttime dynamic scenes is crucial for various applications, including autonomous driving and video surveillance. Conventional cameras often require longer exposure times to capture information in low-light environments, but this inevitably leads to motion blur, which is a significant contradiction for conventional cameras, as shown in Fig.~\ref{fig1_imgaing_example}(a). While existing low-light image/video enhancement techniques can enhance contrast \cite{lore2017llnet,chen2018learning,wang2019underexposed,lv2020fast,zhang2021beyond,zheng2022semantic,chen2019seeing,jiang2019learning,triantafyllidou2020low,zhang2021learning,wang2021seeing,ma2022toward,xu2022snr}, the loss of essential scene structural information leads to suboptimal imaging results. Thus, achieving high-quality imaging in nighttime dynamic scenarios remains an intricate challenge.

Event cameras  \cite{lichtsteiner2008128,posch2010qvga,finateu20205,delbruck2022utility,taverni2018front,suh20201280} overcome the challenge of the trade-off between seeing faster and seeing clearer, especially in nighttime scenarios. This sensor activates each pixel independently in response to changes in brightness and possesses several notable advantages, including high temporal resolution (1$\mu$s) and high dynamic range (120dB), as shown in Fig.~\ref{fig1_imgaing_example}(b). Therefore, reconstructing images from events offers an attractive approach for nighttime imaging.

Early event-based reconstruction methods rely on hand-crafted priors \cite{cook2011interacting,kim2008simultaneous,bardow2016simultaneous,munda2018real}. Recently, learning-based event reconstruction methods \cite{rebecq2019events,wang2019event,stoffregen2020reducing,weng2021event,zou2021learning,cadena2021spade,zhu2022event,liu2023sensing,wang2020eventsr,zhang2020learning,paredes2021back} have made remarkable progress. Sufficient data is the cornerstone of learning methods, but high-quality paired data is often hard to acquire. Unsupervised and self-supervised methods attempt to relax the dependency of paired data through knowledge transfer \cite{wang2020eventsr,zhang2020learning}  or the construction of self-supervised loss functions \cite{paredes2021back}. However, these methods ignore the non-stationary changes of events under complex lighting when constructing the unsupervised learning strategy, resulting in poor performance.

Otherwise, supervised methods \cite{rebecq2019events,stoffregen2020reducing,weng2021event,zhu2022event,liu2023sensing} commonly utilize simulators \cite{rebecq2018esim,hu2021v2e,zhu2021eventgan,lin2022dvs} to generate paired data. While the above simulators provide powerful capabilities, their greatest enemy when applied in real-world scenarios is the diversity and non-uniformity of lighting, especially in the presence of artificial light sources at night. Simulation events tend to have a more uniform signal characteristic, whereas the signal characteristics of events are complex and varied in the real world. To address this issue, many works \cite{han2020neuromorphic,zou2021learning,tulyakov2022time} have set up co-axial imaging systems to collect real-world paired data, but these data were taken in good light and do not provide high-quality GTs for low-light events. Although these state-of-the-art methods have achieved impressive results in well-illuminated conditions, they cannot well generalize to real nighttime environments due to the lack of high-quality paired data.

To cope with the above issue, we first construct a paired real low-light event dataset (RLED). Specifically, we design a co-axial imaging system comprising an event camera with a neutral density(ND) filter and a conventional camera, which allows for the simultaneous acquisition of low-light events and high-quality images. RLED provides 64,200 images with corresponding events, capturing events illuminance levels ranging from 0.5 lux to 1000 lux.

In addition, we propose a nighttime event reconstruction network (NER-Net) including a learnable event timestamps calibration module (LETC) and a non-uniform illumination aware module (NIAM). We first explore the spatiotemporal distribution patterns of events in nighttime lighting conditions based on the sensor circuit characteristics. The sensor exhibits a lower cutoff frequency in low-light conditions, leading to the trailing events effect. As events distribute similarly to the response of an RC low-pass filter \cite{hu2021v2e}, the time interval between adjacent events at the same pixel increases gradually over time. Based on this characterization, we design an event trail suppression (ETS) method that aligns the delayed-trigger events to their correct timestamps and utilizes a LETC to learn the alignment process of events in the time domain, and LETC pre-trained using data preprocessed by ETS. Besides, non-uniform artificial lighting results in significant variations in the distribution of events in different regions. Regions with higher illuminance often have clearer textures and higher event density, and vice versa. In this work, we provide a NIAM to model the non-stationary state of events. NIAM can perform adaptive non-uniform correction based on event density cues and model the non-stationary state of signal distribution by aggregating long-term memory parameters and hierarchical spatial information. As shown in Fig.~\ref{fig1_imgaing_example}(d), NER-Net can effectively reconstruct the intensity of the nighttime dynamic scene. Our main contributions can be summarized as follows:

\begin{itemize}[leftmargin=10pt]

\item We provide an event-based nighttime imaging solution under non-uniform illumination and construct a paired multi-illumination level real-world dataset. To our best knowledge, RLED is the first dataset that provides high-quality pixel-aligned GTs for low-light events.

\setlength{\itemsep}{5pt}
\item We discover the core reasons for the bad performance of existing methods are the temporal trailing effect and the spatial non-uniform response of events, and propose a NER-Net to model the non-stationary status of events under nighttime non-uniform illumination, which effectively improves the reconstruction quality.

\setlength{\itemsep}{5pt}
 \item Extensive experimental results show that our NER-Net outperforms SOTA methods on three real-world nighttime datasets. Besides, the proposed method exhibits the ability to generalize to various lighting scenarios.

\end{itemize}

 \begin{figure*}
   \setlength{\abovecaptionskip}{5pt}
   \setlength{\belowcaptionskip}{-12pt}
   \centering
    \includegraphics[width=0.99\linewidth]{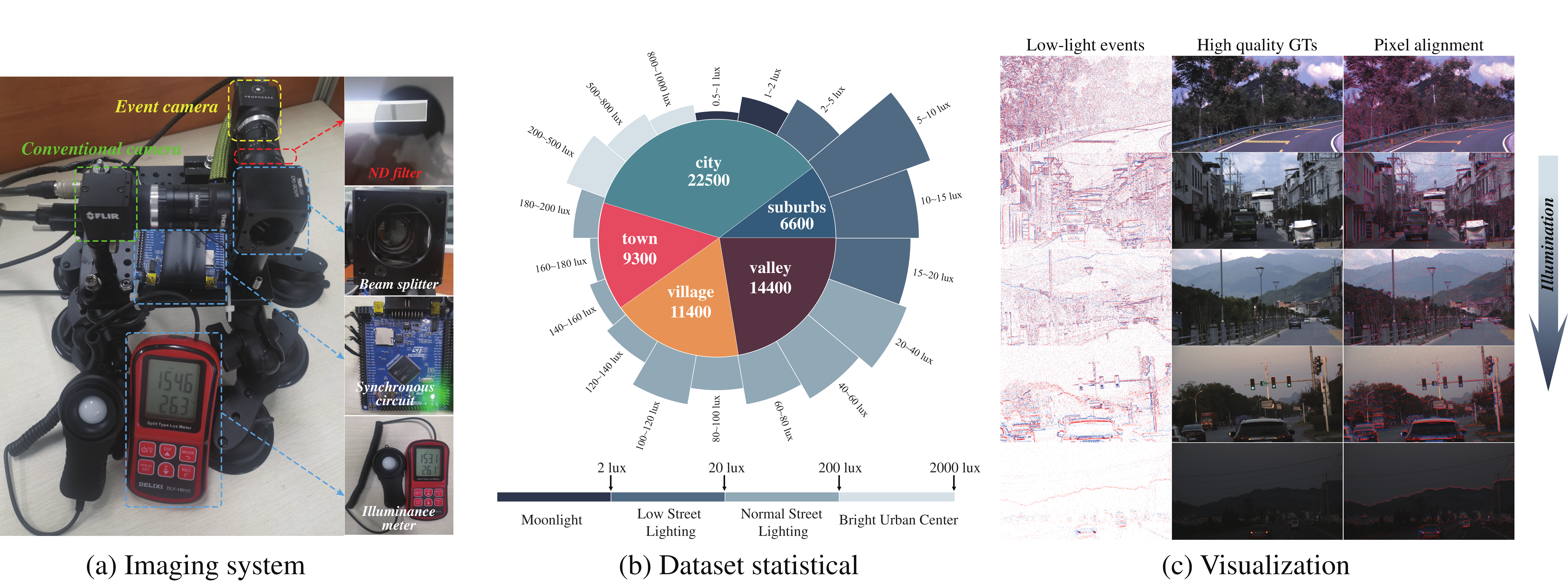}
    \caption{Features of the proposed RLED. (a) The implementation of our coaxial imaging system. (b) Distribution of illumination and scene of the proposed
dataset. (c) Visualization of RLED, which collects low-light events aligned with high-quality images at the pixel level.}
    \label{Dataset}
 \end{figure*}

\section{Related Work}
\label{sec:related_work}

\noindent
\textbf{Nighttime Imaging.}
Frame-based nighttime imaging approaches such as low-light image enhancement (LLIE) have been widely explored. Early LLIE methods were mostly based on histogram equalization \cite{coltuc2006exact,ibrahim2007brightness,arici2009histogram,celik2011contextual} and Retinex theory \cite{jobson1997properties,fu2016weighted,guo2016lime}. In recent years deep learning approaches have achieved remarkable success \cite{lore2017llnet,chen2018learning,wang2019underexposed,lv2020fast,zhang2021beyond,zheng2021adaptive,ma2022toward,mildenhall2022nerf,wu2022uretinex,xu2022snr}. In addition, several works explored infrared nighttime imaging \cite{liu2020deep, luo2022thermal} or hybrid RGB-VIR imaging approaches \cite{ashiba2020hybrid, jia2021llvip}, which enhanced nighttime information acquisition capabilities by broadening the spectral bands. To see moving in the dark, some works \cite{chen2019seeing,jiang2019learning,wang2021seeing} collected paired low-light and day-light video datasets to enhance generalization in real dynamic scenes, and focused on the temporal consistency of videos. Unfortunately, setting a short exposure time for fast-moving scenes leads to significant loss of scene details. Frame-based methods cannot recover lost information.

Recently, nighttime imaging based on events became a hot topic. DVS-Dark \cite{zhang2020learning} learns common features (scene structures) between the day and night and transfers daytime domain-specific knowledge (detailed textures) to the nighttime domain. Furthermore, event-guided image/video enhancement approaches have been explored for nighttime imaging \cite{liu2023low,liang2023coherent,jiang2023event,wang2023event}. These works skillfully leveraged the high dynamic range and high speed of event cameras to help images recover scene intensity. However, it is worth noting that the spatiotemporal distribution of events in real nighttime scenes is complex and variable. Neglecting these factors can significantly impact the quality of imaging. In this work, we analyze the spatiotemporal distribution patterns of events in nighttime lighting based on the sensor circuit characteristic and design a NER-Net to address the temporal trailing effect and spatial non-uniform distribution of events.

\noindent
\textbf{Event-based image and video reconstruction.}
Recovering high speed and high dynamic range intensity images from events is an efficient imaging approach. Early methods attempted to construct mathematical models of event generation or introduced artificial priors to recover image intensity \cite{cook2011interacting,kim2008simultaneous,bardow2016simultaneous,munda2018real}, but suffered from severe artifacts or over-smoothed results. With the development of deep learning technology, the quality of reconstructed images has seen an impressive improvement \cite{rebecq2019events,wang2019event,stoffregen2020reducing,weng2021event,zou2021learning,cadena2021spade,zhu2022event,liu2023sensing,wang2020eventsr,zhang2020learning,paredes2021back}. Rebecq \etal \cite{rebecq2019events} proposed an effective recurrent network architecture and leveraged a large amount of simulated data to learn the mapping from events to images. However, the sensor noise can lead to serious reconstruction artifacts due to the limited diversity of simulated data. Thus, Stoffregen \etal \cite{stoffregen2020reducing} and Liu \etal \cite{liu2023sensing} attempt to improve generalization in real-world scenarios by enhancing the simulation process. In contrast, Zou \etal \cite{zou2021learning} collected a paired events and images dataset through a co-axial imaging system, and the HDR ground truth was synthesized by combining two low dynamic range (LDR) images with different brightness levels. Unfortunately, this method cannot generate high-quality ground truth at night.

Although these methods have achieved impressive results in well-illuminated scenes, the lack of low-light paired data makes them ineffective in real nighttime scenes. In this work, we contribute a multi-illumination paired dataset, which provides high-quality pixel-aligned GTs for low-light events.

\section{Nighttime Event Signal Characteristics}
\label{sec:analysis}

As mentioned above, paired low-light event and high-quality image datasets are still lacking, and the spatiotemporal non-stationary characteristics of nighttime events remain under-explored. In this section, we first present the real low-light event dataset (RLED). Then, we analyze the response properties of events in the temporal domain and point out that the core reason for the event trailing effect is the reduced sensor cutoff frequency. Furthermore, we discuss the impact of non-uniform artificial lighting on the spatial distribution of events, higher illumination near the light source typically leads to higher event density, and vice versa.

\subsection{Real Low-light Event Dataset}
Obtaining high-quality paired data in nighttime dynamic scenes is virtually an impossible task because conventional cameras have poorer imaging capabilities at night. To address this issue, we collect data under optimal lighting conditions during the daytime, while also equipping the event camera with an ND filter to obtain low-light events.

First, we built a co-optical axis imaging system comprising an event camera (Prophesee EVK4, 1280*720), a conventional camera (FLIR BFS-U3-32S4C, 2048*1536), a beam splitter (Thorlabs BSW26R) and a ND filter (Thorlabs ND20A), as illustrated in Fig.~\ref{Dataset}(a). The beam splitter divides the incoming light into two equal parts and sends them to the event camera and the conventional camera respectively. The ND filter is placed between the event camera and the beam splitter to filter out the majority of incoming light (99\%). In addition, we provided external trigger signals to the cameras through a programmable synchronous circuit, enabling precise synchronization of the timestamps of both cameras. Finally, we achieved pixel alignment between the two cameras through the stereo rectification process.

RLED contains 64,200 images and corresponding events, we utilized a photometer to continuously measure scene illumination, and calculate the illuminance value after attenuation at the event camera.  As shown in Fig.~\ref{Dataset}(b), the illumination range for the event data was between 0.5 lux to 1000 lux. The capture scenes included city (35.0\%), suburbs (10.3\%), town (14.5\%), village (17.8\%), and valley (22.4\%).

 \begin{figure}
   \setlength{\abovecaptionskip}{5pt}
   \setlength{\belowcaptionskip}{-10pt}
   \centering
    \includegraphics[width=1.0\linewidth]{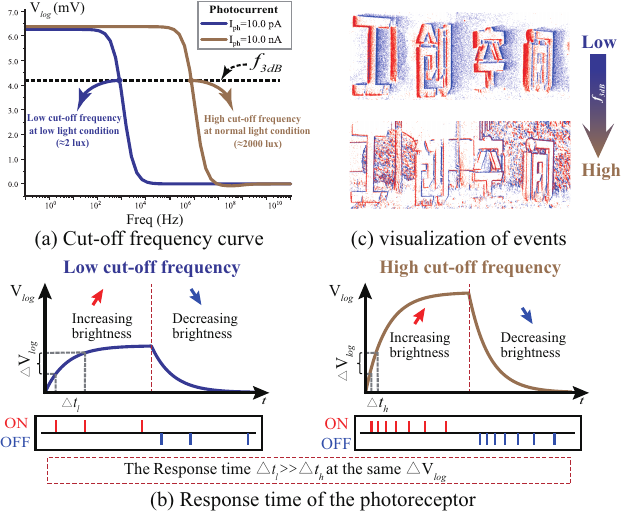}
    \caption{This figure illustrates the mechanism of trailing events. (a) Photoreceptor cut-off frequency decreases with decreasing illumination. (b) The response time of the photoreceptor increases as the cut-off frequency decreases, requiring more time($\Delta t_l$ \textgreater $\Delta t_h$) to reach the same voltage change $V_{log}$ at low illumination. (c) The increase in response time leads to the effect of trailing events.}
    \label{Trailing events}
 \end{figure}

\subsection{Trailing Events in Temporal Domain}
The signal characteristics of event cameras can be influenced by external factors such as temperature, notably illumination. The photocurrent of the photodiode $I_{ph}$ is directly related to the intensity of illumination,
\begin{equation}\
  \setlength\abovedisplayskip{3pt}
  \setlength\belowdisplayskip{3pt}
  \begin{aligned}
   I_{ph} = \textit{R}  \times \textit{A}\times \textit{E},
 \label{eq:photocurrent}
 \end{aligned}
\end{equation}
where \textit{R} is the responsivity of the photodiode, \textit{A} is the photodiode active area, \textit{E} is the illuminance of the incident light. The magnitude of photocurrent directly influences cutoff frequency $f_{3dB}$ of the sensor, which is a key indicator used to evaluate signal response speed. The magnitude of the cutoff frequency is directly correlated with the photocurrent \cite{posch2010qvga},
\begin{equation}\
  \setlength\abovedisplayskip{3pt}
  \setlength\belowdisplayskip{3pt}
  \begin{aligned}
   \resizebox{0.4\hsize}{!}{$f_{3dB} = \frac{1}{2\pi C_{\rm DMfb}} \frac{I_{ph}}{U_t}$},
 \label{eq:f3dB}
 \end{aligned}
\end{equation}
where $C_{\rm DMfb}$ is the gate-drain capacitance of transistor $M_{\rm fb}$ and $U_t$ is the thermal voltage. When the temperature is constant, $f_{3db}$ is proportional to $I_{ph}$. Therefore, when the environmental illumination changes, it directly affects the cutoff frequency of the sensor, thereby altering the signal characteristics of events.

The variation of the cutoff frequency with changes in illumination is depicted in Fig.~\ref{Trailing events}(a). The cutoff frequency represents the ability of the photodiode voltage $V_{log}$ to track changes in photocurrent $I_{ph}$. When the cutoff frequency is high, the voltage can rapidly respond to current changes, leading to the quick generation of events, and vice versa. As shown in Fig.~\ref{Trailing events}(b), when the cutoff frequency is lower, it requires more response time for the same voltage change $\Delta V_{log}$. This behavior is analogous to the response process of a first-order RC low-pass filter \cite{hu2021v2e}, where the intervals between consecutive events progressively increase. Fig.~\ref{Trailing events}(c) illustrates the trailing events resulting from the lower cutoff frequency of the sensor under low light conditions, which significantly blurs the edges of the scene.

 \begin{figure}
   \setlength{\abovecaptionskip}{5pt}
   \setlength{\belowcaptionskip}{-7pt}
   \centering
    \includegraphics[width=1.0\linewidth]{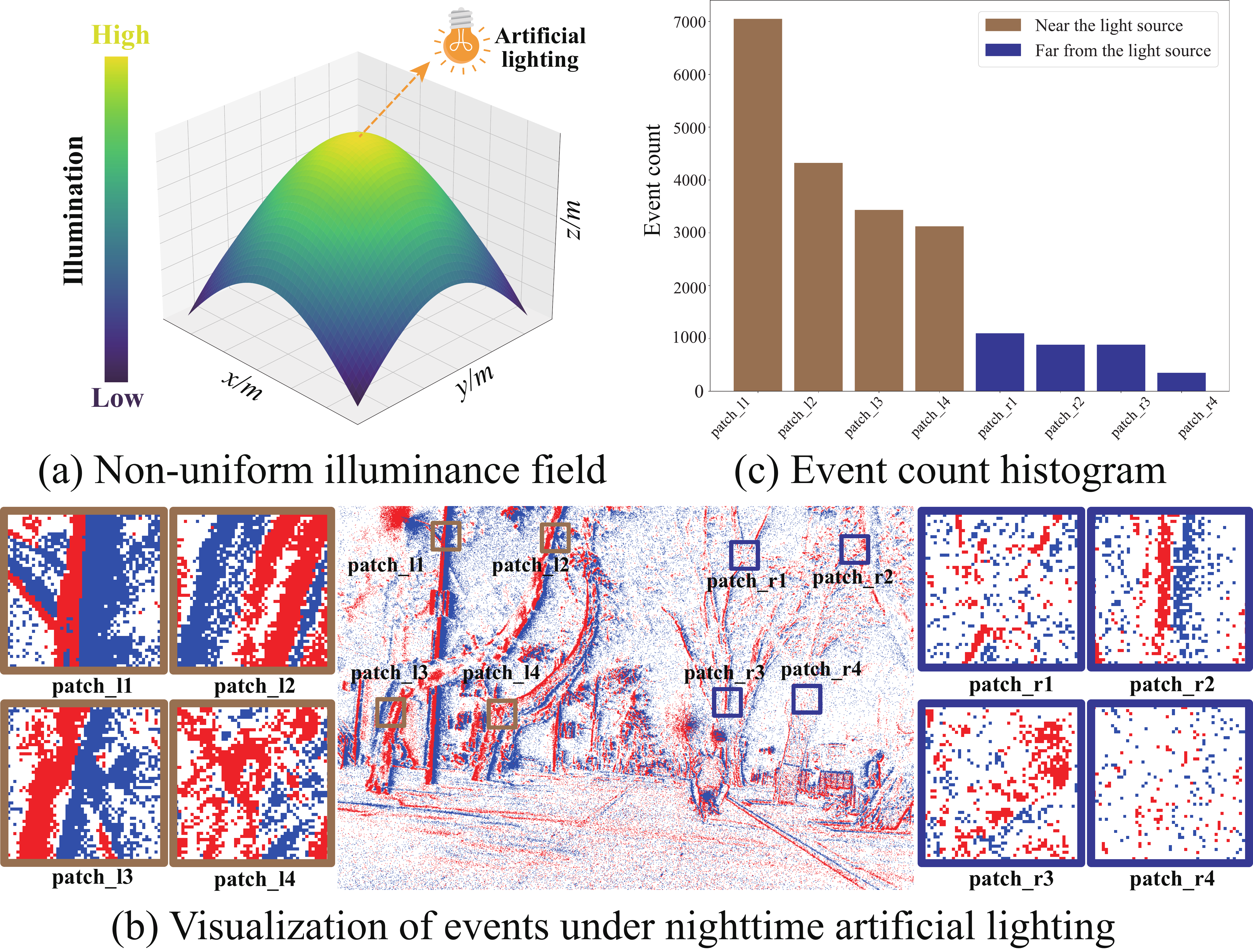}
    \caption{The spatial characteristics of events at night. (a) The illuminance field under artificial nighttime lighting is non-uniform and exhibits significant variations with different distances from the light source. (b) Events are more densely distributed in the vicinity of the artificial light source. (c) Greater luminance difference between nearer the light source and the dark background, resulting in more events being triggered, and vice versa.
    }
    \label{nom-uniform_Illumination}
 \end{figure}

 \begin{figure*}
   \setlength{\abovecaptionskip}{5pt}
   \setlength{\belowcaptionskip}{-5pt}
   \centering
    \includegraphics[width=0.95\linewidth]{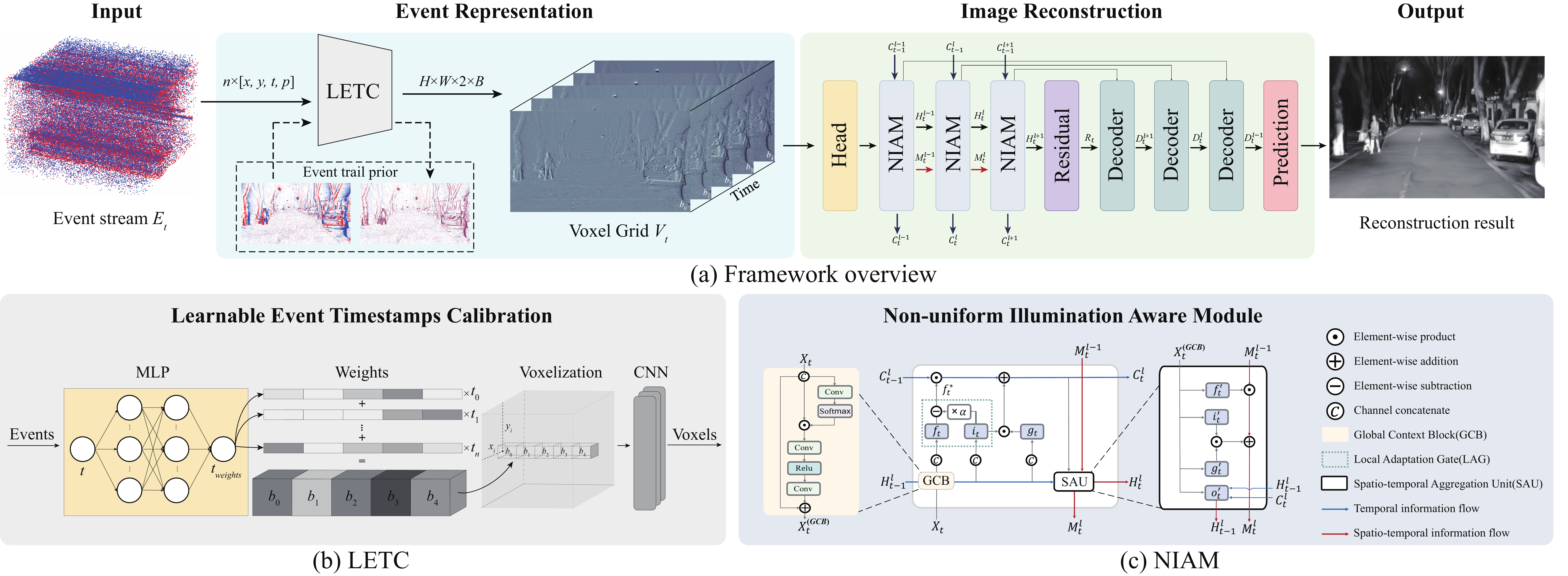}
    \caption{The overall architecture of the proposed \textbf{Nighttime Event Reconstruction(NER)} network, which can model the non-stationary spatiotemporal distribution of nighttime events.
(a) NER contains a Learnable Event Timestamps Calibration (LETC) and a U-shaped image reconstruction network with Non-uniform Illumination Aware Module (NIAM) encoders. (b) LETC generates voxels with sharp edges by redistributing the weights of event timestamps within different voxel units. (c) NIAM performs local and global illumination sensing and regulation via the Local Adaptation Gate(LAG) and Global Context Block(GCB), respectively. By using a Spatiotemporal Aggregation Unit(SAU), NIAM can adaptively exploit and fuse multi-scale spatial information and long-term temporal information.}
    \label{framework}
 \end{figure*}

 \subsection{Non-uniform Illumination in Spatial Domain}
In nighttime autonomous driving or surveillance, artificial lighting sources such as street lamps are often present. One significant characteristic of artificial light sources is their non-uniformity. The illuminance formula is
\begin{equation}\
  \setlength\abovedisplayskip{3pt}
  \setlength\belowdisplayskip{3pt}
  \begin{aligned}
   I = \frac{\Phi}{4\pi d^2},
 \label{eq:illu_dist}
 \end{aligned}
\end{equation}
where $I$ is the illuminance in the scene, $\Phi$ is the luminous flux, which depends on the power of the light source, and $d$ is the distance from the light source to the scene. The power of the lighting can be regarded as constant, then the only factor affecting the illuminance in the scene is the distance from the light source. The illuminance field of the artificial light source is shown in Fig. \ref{nom-uniform_Illumination}(b), the closer the distance to the light source, the higher the illuminance.

The event generation process \cite{paredes2021back} can be formulated as
\begin{equation}\
  \setlength\abovedisplayskip{3pt}
  \setlength\belowdisplayskip{3pt}
  \begin{aligned}
   logI(x, y, t) - logI(x, y, t - \Delta t) = p C,
 \label{eq:illu_dist}
 \end{aligned}
\end{equation}
where $logI(x, y, t)$ is the logarithmic illuminance at pixel $(x, y)$ and time $t$, $\Delta t$ is the time interval between consecutive events, $p\in[-1, 1]$ is the polarity of events, and $C$ is the contrast threshold of the event camera. This process indicates that an event is triggered once the logarithmic illuminance change at a particular pixel exceeds the threshold $C$. Due to the extremely low background brightness, the closer the distance to the light source, the greater the difference in brightness between the object and the background, leading to the triggering of more events. As shown in Fig. \ref{nom-uniform_Illumination}(a), the distance to the light source is closer on the left side, and there is a significantly higher event density compared to the right side. The lower illumination on the right side results in a lower event density and weak texture as in Fig. \ref{nom-uniform_Illumination}(c).


 \section{Nighttime Event Reconstruction Network}
 Fig. \ref{framework} illustrates the overall framework of nighttime event reconstruction network (NER-Net).

\subsection{Trail Prior Guided Event Representation}

 \begin{figure}
   \setlength{\abovecaptionskip}{3pt}
   \setlength{\belowcaptionskip}{-15pt}
   \centering
    \includegraphics[width=1.0\linewidth]{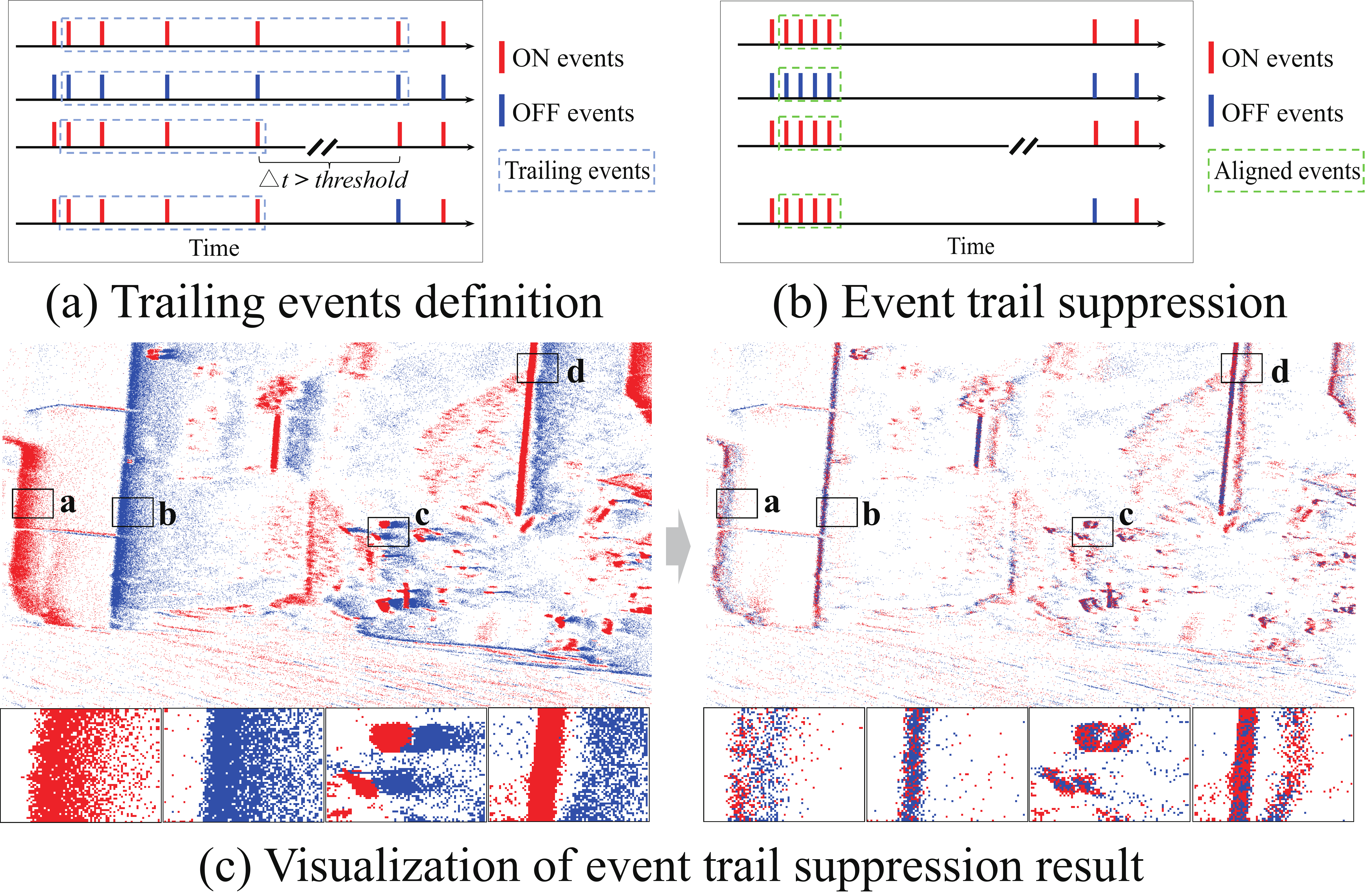}
    \caption{Method and visualization of Event Trail Suppression (ETS). (a) ETS is designed based on the event response characteristics in Section \textcolor{red}{3.2} with three conditions. (b) ETS aligns event timestamps to the correct position.
    (c) Events processed by ETS result in sharper edges.}
    \label{ETS}
 \end{figure}

 \begin{figure*}
   \setlength{\abovecaptionskip}{5pt}
   \setlength{\belowcaptionskip}{-10pt}
   \centering
    \includegraphics[width=0.96\linewidth]{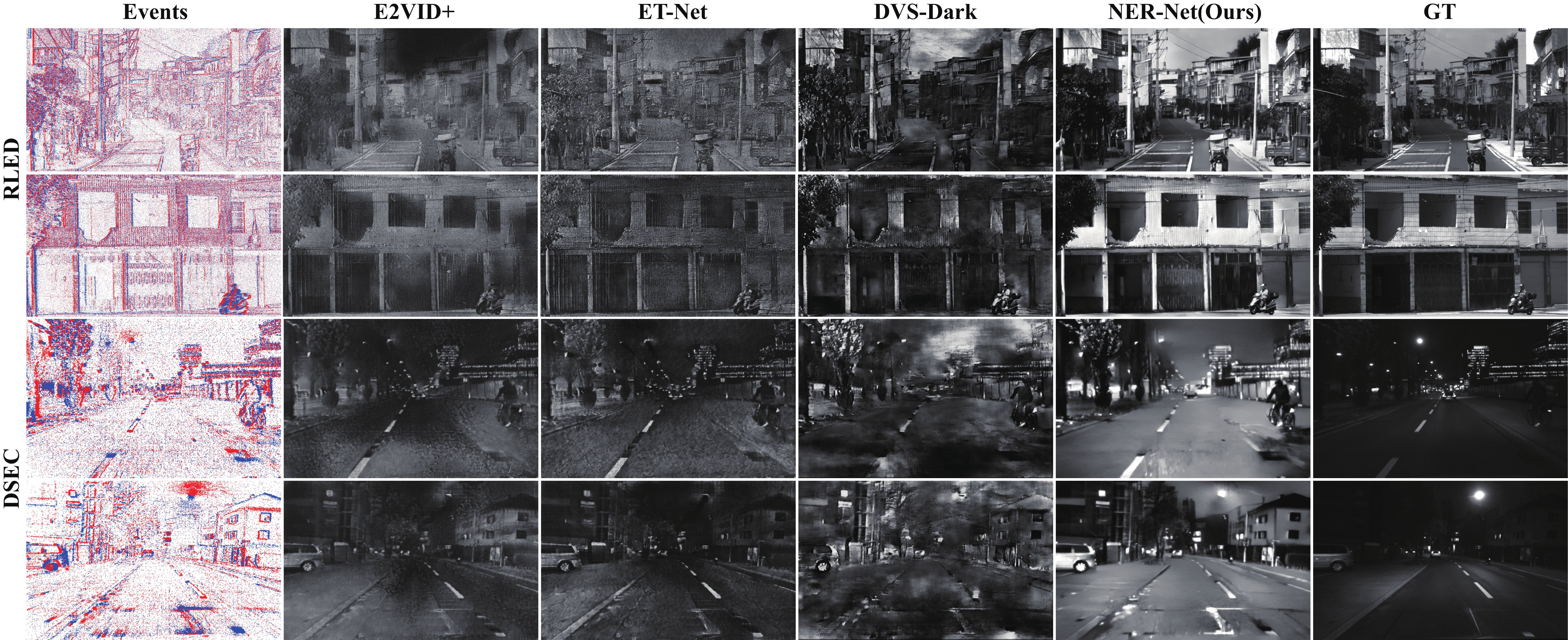}
    \caption{Visual comparison between other SOTA methods and proposed NER-Net on real-world datasets.}
    \label{qualitative_results}
 \end{figure*}

\noindent
\textbf{Event Trail Suppression.}
In Section \ref{sec:analysis}, we discuss the response of events is similar to an RC low-pass filter, the rate of change in photocurrent voltage gradually decreases after an excitation. This phenomenon becomes more pronounced with lower cutoff frequencies, resulting in progressively longer time intervals between consecutive events. We design the ETS method based on this characteristic, and the conditions for identifying tailing events include: (1) the polarity of consecutive events remains unchanged,  (2) the time intervals between consecutive events are gradually increasing, (3) the largest time interval among consecutive events is less than a specified threshold, as shown in Fig.~\ref{ETS}(a).

Once a trailing event has been identified, ETS will correct the timestamp of the event. The corrected timestamps are rearranged at a specified interval, which is specified to be 1$\mu$s based on statistics in daylight, as shown in Fig.~\ref{ETS}(b). Fig.~\ref{ETS}(c) illustrates the visual results of ETS.

\noindent
\textbf{Learnable Event Timestamps Calibration.}
To enhance the reconstruction quality with the prior of tail suppression, we design an end-to-end trainable LETC module. The LETC can transform the trailing events to sharp voxel grids by assigning weights to timestamps. The classic voxelization method transform events $\varepsilon_{k}=\left \{ e_i \right \} _{i=0}^{N-1}$ into a tensor $V\in \mathbb{R}^{B\times H \times W}$ with $B$ bins \cite{zhu2019unsupervised}, which can be formulated as

\begin{equation}\
  \setlength\abovedisplayskip{-5pt}
  \setlength\belowdisplayskip{5pt}
  \begin{aligned}
   \resizebox{0.8\hsize}{!}{$V(k)=\sum_ip_i\max(0, 1-\left |k- \frac{t_i-t_0}{t_N-t_0}(B-1) \right|)$},
 \label{eq:voxel}
 \end{aligned}
\end{equation}
where $N$ is the number of events, $p_i$ and $t_i$ represent the polarity and timestamp of the $i$-th event respectively, and $k \in [0, B - 1]$. This method uniformly fills events into the two nearest bins using the fixed interpolation approach.

In order to make the event representation process more adaptable to specific tasks, Gehrig \etal \cite{gehrig2019end} introduced a representation method called event spike tensor (EST) that allows each event to be filled into all B bins, and the filling values are learned end-to-end by an MLP. Motivated by this, the proposed LETC module can dynamically assign weights based on the degree of trailing events to generate sharp voxels dynamically allocates weights based on the degree of event tailing, thereby generating sharp voxels. Differing from their approach of directly embedding the representation module into the task network for joint training \cite{gehrig2019end}, the LETC module includes a tail CNN layer to integrate information for voxel generation, and then it uses ETS-processed events as labels to train LETC, enabling LETC to have a priori knowledge for trailing suppression. Finally, LETC is integrated into the reconstruction network for fine-tuning.

\subsection{Non-uniform Illumination Aware Module}
Events are temporally continuous and spatially sparse, the restoration of scene intensity relies on modeling the continuous spatiotemporal context. Non-uniform illumination disrupts the spatiotemporal distribution stability of events, leading to significant reconstruction artifacts. Hence, stabilizing the spatiotemporal distribution of events is crucial for restoring scene intensity. Inspired by the human vision system (HVS), which can globally regulate light intake through the expansion and contraction of the pupil, retinal cells can locally adjust based on light intensity through the response compression mechanism\cite{crawford1947visual,boff1986handbook}. We design NIAM encoders comprising GCB, LAG, and SAU modules.

GCB \cite{cao2019gcnet} captures long-range dependencies in non-uniform illumination fields, which can be formulated as
\begin{equation}\
  \setlength\abovedisplayskip{3pt}
  \setlength\belowdisplayskip{3pt}
  \begin{aligned}
   \resizebox{0.9\hsize}{!}{$\mathcal{X} = x + Conv_{1\times 1} (ReLU (LN(Conv_{1\times 1}(\sum_{j=1}^{N}\gamma _jx_j ))))$,}
 \label{eq:GCB}
 \end{aligned}
\end{equation}
where $N$ is the number of positions in the feature map, $\gamma_j=\frac{e^{Conv_{1\times 1}(x_j)}}{\sum_{m}e^{Conv_{1\times 1}(x_m)}}$ is the weight for global attention pooling.

\noindent
\textbf{Local Adaptation Gate.}
The standard convLSTM \cite{shi2015convolutional} can effectively handle temporal information. Still, we aim to preserve weak texture area while eliminating redundant information in the temporal memory parameter. The event density encoded information is additionally introduced in the LAG and acts as negative feedback to the forgot gate. This allows for adaptive control of the memory parameters based on the density of input events. The architecture of LAG is illustrated in the green box of Fig.~\ref{framework}(c), formulated as
\begin{equation}\
  \setlength\abovedisplayskip{3pt}
  \setlength\belowdisplayskip{3pt}
  \begin{aligned}
    f_t^{LAG} =  \sigma (f_t - \alpha i_t),
 \label{eq:GCB}
 \end{aligned}
\end{equation}
where $\alpha = e^{\sigma (Conv(x_t))}$ is the event density encoded information, $\sigma(\cdot )$ denotes sigmoid function, $f_t$ and $i_t$ represent the forgot gate and input gate of convLSTM respectively. Note that, event density depends on both illumination and object motion. LAG regulates information flow based on event density without requiring their decoupling. The effectiveness of the proposed method is shown in Fig.~\ref{qualitative_results}.

\noindent
\textbf{Spatiotemporal Aggregation Unit.}
ConvLSTM-based encoder are widely used in event-based image reconstruction networks\cite{rebecq2019events,stoffregen2020reducing,weng2021event,cadena2021spade,zou2021learning,paredes2021back}, but they focus more on modeling temporal variations and memory states being updated repeatedly over time inside each LSTM unit, resulting in information blockage between different layers. Inspired by PredRNN \cite{wang2017predrnn}, we design the SAU to integrates long-term temporal information along with hierarchical spatial information, as shown within the black dashed box Fig. \ref{framework}(c). $\mathcal{C}_t^l$ is the temporal cell including GAB and LAG, which being updated repeatedly over time. $\mathcal{M}_t^l$ is the spatiotemporal memory that can aggregate hierarchical features across layers, and it also transmit the spatiotemporal memory from the top layer at time $t-1$ with rich semantic information to the bottom layer at time $t$ through a progressive upsampling. Thus, NIAM can effectively perceive the non-stationary state of signal distribution, and model the dynamic information in non-uniform illumination environments. The detailed network architecture can be found in supplementary material.

\subsection{Training Details}
\noindent
\textbf{Loss Function.}
We use LPIPS \cite{zhang2018unreasonable} to evaluate image quality, and temporal consistency(TC) loss \cite{lai2018learning}  is employed to mitigate the temporal artifacts between adjacent reconstructed images. The total loss is
\begin{equation}\
  \setlength\abovedisplayskip{3pt}
  \setlength\belowdisplayskip{3pt}
  \begin{aligned}
     {\textstyle \sum_{i=0}^{L}} \mathcal{L}_i^{LPIPS} + \lambda_{TC} {\textstyle \sum_{i=L_0}^{L}} \mathcal{L}_i^{TC},
 \label{eq:Loss}
 \end{aligned}
\end{equation}
where $L$ and $L_0$  are the iteration time step and the starting index for TC loss calculations. We set $L = 40$, $L_0 = 2$ , and $\lambda_{TC} = 2$ respectively.

\noindent
\textbf{Implementation.}
Our network is implemented using the Pytorch framework \cite{paszke2017automatic} and use
ADAM \cite{kingma2014adam} with a learning rate of 0.0001. Patches at the size of 160×160 are randomly cropped from training samples. We train our model for 200 epochs using an NVIDIA A100 GPU.


\begin{table}
\footnotesize
  \renewcommand\arraystretch{1.1}
  \centering
  \setlength\tabcolsep{3pt}
  \setlength{\abovecaptionskip}{5pt}
  \setlength{\belowcaptionskip}{-12pt}

\resizebox{0.5\textwidth}{!}{
\begin{tabular}{crccccccccccc}
	\toprule[1pt]
    \multirow{2}{*}{Datasets}& Methods & \multicolumn{3}{c}{E2VID+} & \multicolumn{3}{c}{ET-Net} & DVS-Dark & \multicolumn{3}{c}{\textbf{NER-Net(ours)}} \\
    \cmidrule(r){2-2} \cmidrule(r){3-5} \cmidrule(r){6-8} \cmidrule(r){9-9} \cmidrule(r){10-12}

    & {\makecell[c]{Training \\ data}}  & $^*$ESIM & V2E & \textbf{RLED}  & $^*$ESIM & V2E & \textbf{RLED} & $^*$DVS-Dark & ESIM & V2E & \textbf{RLED} \\
    \midrule[1pt]

    \multirow{3}{*}{RLED} & MSE $\downarrow$ &0.099 &0.100 &\textcolor[RGB]{5, 119, 72}{\textbf{0.019}} &0.051 &0.096 &0.075 &0.095 &0.077 & 0.087 & \textcolor[RGB]{0, 52, 114}{\textbf{0.011}} \\
    & SSIM $\uparrow$ &0.298 &0.404 &\textcolor[RGB]{5, 119, 72}{\textbf{0.688}} &0.307 &0.421 &0.534 &0.355 &0.179 &0.081 & \textcolor[RGB]{0, 52, 114}{\textbf{0.717}} \\
    & LPIPS $\downarrow$ &0.560 &0.582 &\textcolor[RGB]{5, 119, 72}{\textbf{0.393}} &0.559 &0.544 &0.411 &0.591 &0.628 &0.749  & \textcolor[RGB]{0, 52, 114}{\textbf{0.309}} \\  \hline

    \multirow{3}{*}{{\makecell[c]{DSEC \\ -night}}} & LOE $\downarrow$ &1383.4 &1704.2 &\textcolor[RGB]{5, 119, 72}{\textbf{1165.1}} &1271.1 &1782.9 &1227.6 &1757.7 &1421.2 &1694.9 & \textcolor[RGB]{0, 52, 114}{\textbf{1031.2}} \\
    & SSIM $\uparrow$ &0.311 &0.304 &0.323 &0.331 &0.272 &\textcolor[RGB]{0, 52, 114}{\textbf{0.411}} &0.337 &0.153 & 0.305 & \textcolor[RGB]{5, 119, 72}{\textbf{0.340}} \\
    & LPIPS $\downarrow$ &0.586 &0.554 &\textcolor[RGB]{5, 119, 72}{\textbf{0.534}} &0.581 &0.552 &\textcolor[RGB]{5, 119, 72}{\textbf{0.534}} &0.608 &0.622 & 0.536 & \textcolor[RGB]{0, 52, 114}{\textbf{0.502}} \\  \hline

    \multirow{3}{*}{{\makecell[c]{MVSEC \\ -night}}} & LOE $\downarrow$ &1292.9 &1708.5 &\textcolor[RGB]{5, 119, 72}{\textbf{1283.8}} &1588.9 &1739.7 &1344.3 &1604.2 &1534.1 &1743.7 & \textcolor[RGB]{0, 52, 114}{\textbf{1198.4}} \\
    & SSIM $\uparrow$ &0.078 &0.104 &0.141 &0.113 &0.085 &\textcolor[RGB]{5, 119, 72}{\textbf{0.197}} &0.120 &0.062 &0.092 & \textcolor[RGB]{0, 52, 114}{\textbf{0.205}} \\
    & LPIPS $\downarrow$ &0.579 &0.644 &0.564 &0.573 &0.658 &\textcolor[RGB]{5, 119, 72}{\textbf{0.507}} &0.598 &0.606 &0.645 & \textcolor[RGB]{0, 52, 114}{\textbf{0.482}} \\ \hline

    \multirow{3}{*}{{\makecell[c]{VECtor \\ -hdr}}} & LOE $\downarrow$ &1495.0 &1824.8 &\textcolor[RGB]{5, 119, 72}{\textbf{1384.5}} &1617.9 &1932.4 &1526.5 &1759.2 &1439.8 &1839.8 & \textcolor[RGB]{0, 52, 114}{\textbf{1139.8}} \\
    & SSIM $\uparrow$ &0.224 &0.184 &0.269 &0.208 &0.153 &0.232 &\textcolor[RGB]{5, 119, 72}{\textbf{0.317}} &0.197 &0.219 & \textcolor[RGB]{0, 52, 114}{\textbf{0.344}} \\
    & LPIPS $\downarrow$ &0.695 &0.707 &\textcolor[RGB]{5, 119, 72}{\textbf{0.652}} &0.701 &0.711 &0.683 &0.723 &0.658 &0.682 & \textcolor[RGB]{0, 52, 114}{\textbf{0.634}} \\

	\bottomrule[1pt]

    \end{tabular}
    }
	\begin{tablenotes}
    \footnotesize
    \raggedright
	\item[*] $^*$ Using the pretrained model provided in the original paper. \\
    \end{tablenotes}
  \caption{Quantitative comparisons with SOTA methods on three real-world datasets: RELD, DSEC-night, and MVSEC-night.  The top two results are colored in \textcolor[RGB]{0, 52, 114}{\textbf{blue}} and \textcolor[RGB]{5, 119, 72}{\textbf{green}}.}
 \label{tab:quantitative evaluation}
\end{table}

\section{Experiments}
\subsection{Datasets and Experimental Settings}
\noindent
\textbf{Datasets.}
Since there is a gap between the simulated data and the real world data, our target is to assess the applicability of methods in real scenarios. Thus, we choose four real low-light datasets to validate the performance of our method: RLED, DSEC \cite{gehrig2021dsec} MVSEC \cite{zhu2018multivehicle} and VECtor \cite{gao2022vector}.

\noindent
\textbf{Comparison Methods.}
We compare our models with three SOTA  methods E2VID+ \cite{stoffregen2020reducing}, ET-Net \cite{weng2021event}, and DVS-Dark \cite{zhang2020learning}. To ensure the fairness of the experiments, in addition to using the pre-trained model provided in the original paper, we re-train the above model using paired data from RLED and V2E \cite{hu2021v2e} simulated data based on RLED images using default noisy settings.

\noindent
\textbf{Evaluation Metrics.}
 For RLED, we use mean squared error (MSE), structural similarity (SSIM) \cite{wang2004image}, and perceptual similarity (LPIPS) \cite{zhang2018unreasonable} as evaluation metrics. For the other three datasets, considering the low quality of nighttime images, we opt for a no-reference image quality assessment metric known as lightness-order-error (LOE) \cite{wang2013naturalness} instead of MSE. LOE effectively measures the naturalness of images in non-uniform illumination conditions.

\subsection{Quantitative and Qualitative Evaluation}
\noindent
\textbf{Qualitative Evaluation.}
Fig. \ref{qualitative_results} depicts a visual comparison of results between other SOTA methods and proposed NER-Net in low-light dynamic scenes.  E2VID+ \cite{stoffregen2020reducing} and ET-Net \cite{weng2021event} exhibit temporal memory errors in the presence of noise and artificial light sources, leading to severe black artifacts. DVS-Dark \cite{zhang2020learning} is a GAN-based method that experiences model breakdowns when encountering diverse data distributions, resulting in unnatural reconstruction results. NER-Net achieves more visually pleasing high dynamic range results in non-uniform illumination and high-noise environments.

 \begin{figure}
   \setlength{\abovecaptionskip}{5pt}
   \setlength{\belowcaptionskip}{-15pt}
   \centering
    \includegraphics[width=0.90\linewidth]{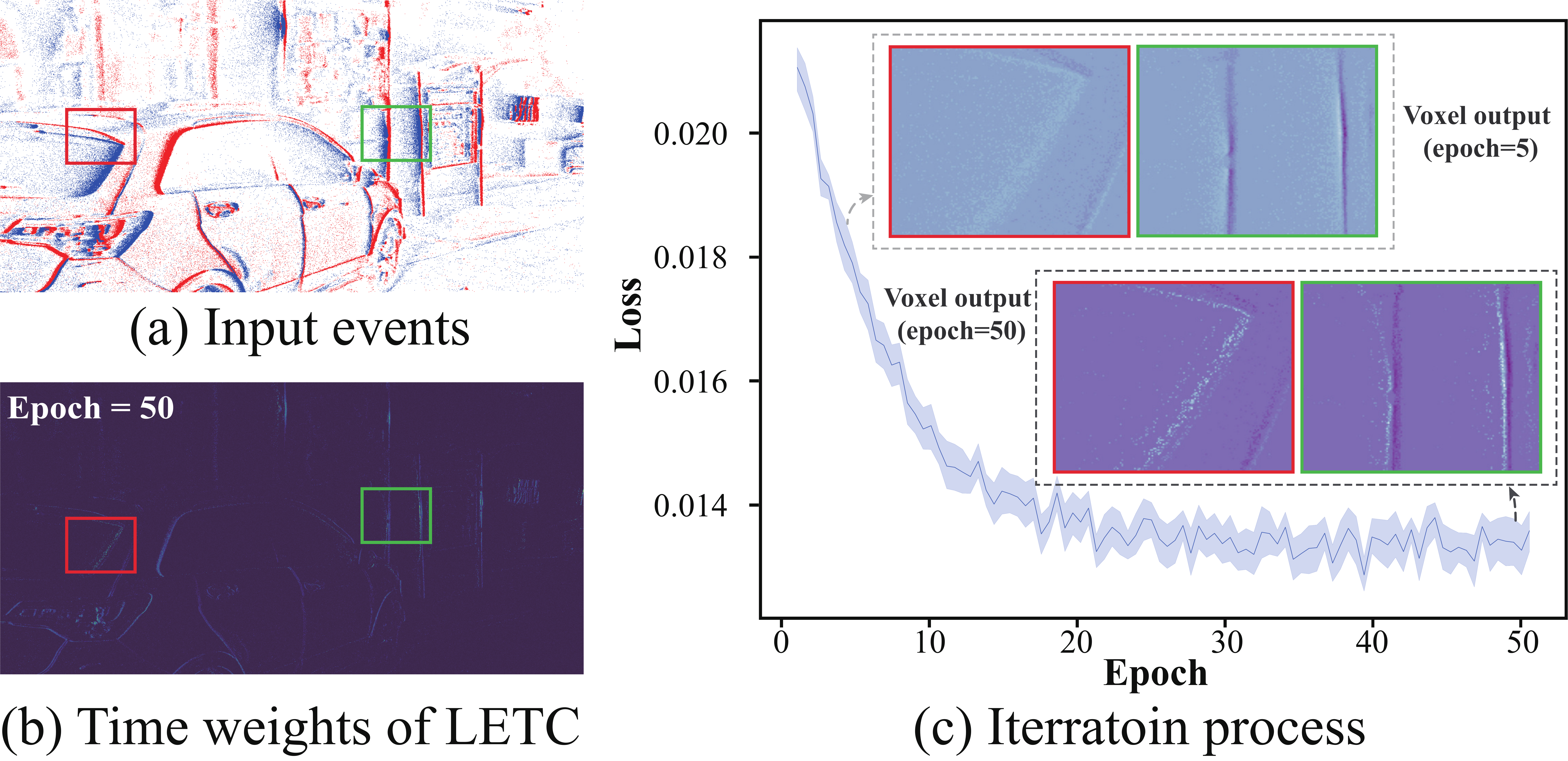}
    \caption{Effectiveness of LETC. (a) Input events. (b) Visualization of $t_{weights}$ at the 50th epoch.
    (c) LETC can generate sharp voxels from trailing events by training with ETS data.}
    \label{ablation_LETC}
 \end{figure}

\noindent
\textbf{Quantitative Evaluation.}
The quantitative results are reported in Table~\ref{tab:quantitative evaluation}. It is observed that all the top two results were trained on RLED. Note that, models trained on a single noise setting using V2E \cite{hu2021v2e} often exhibit poor performance. The reason is that the spatiotemporal distribution of nighttime events is non-uniform and dynamically changing, and there is a large gap with simulated events. NER-Net outperforms the SOTA methods on all three benchmarks, which confirms the effectiveness of our method with event timestamps calibration and non-uniform illumination awareness. More importantly, NER-Net achieved better generalization across four datasets captured by different sensors. We also validate the generalizability of the proposed method during the day. Please refer to supplementary material for details.

\subsection{Ablation Study and Discussion}
\label{sec:ablation}

\begin{table}
\footnotesize
  \centering
  \setlength{\abovecaptionskip}{5pt}
  \setlength{\belowcaptionskip}{-10pt}
  \resizebox{0.48\textwidth}{!}{
  \begin{tabular}{ccccc|ccc}
    \Xhline{1px}
	\multicolumn{2}{c}{LETC} & \multicolumn{3}{c}{NIAM} \vline &  \multicolumn{3}{c}{RLED} \\
    \hline
     ETS & LETC & LAG & GCB & SAU & MSE $\downarrow$ & SSIM $\uparrow$ & LPIPS $\downarrow$ \\
     \Xhline{1px}
     & & & & &0.019 &0.688 &0.393 \\
    \checkmark & & & & &0.045 &0.586 &0.423 \\
    &\checkmark & & & &0.017 &0.692 &0.368 \\
    \checkmark &\checkmark & & & &0.016 &0.704 &0.354 \\ \hline
     \checkmark &\checkmark &\checkmark & & &0.014 &0.710 &0.331 \\
     \checkmark &\checkmark &\checkmark &\checkmark & &0.015 &0.712 &0.324 \\
    \checkmark &\checkmark &\checkmark &\checkmark &\checkmark &\textbf{0.011} &\textbf{0.717} &\textbf{0.309} \\
    \Xhline{1px}
    \end{tabular}}
    \caption{Ablation studies of NER-Net.}
 \label{tab:ablation_network}
\end{table}

 \begin{figure}
   \setlength{\abovecaptionskip}{5pt}
   \setlength{\belowcaptionskip}{-15pt}
   \centering
    \includegraphics[width=1.0\linewidth]{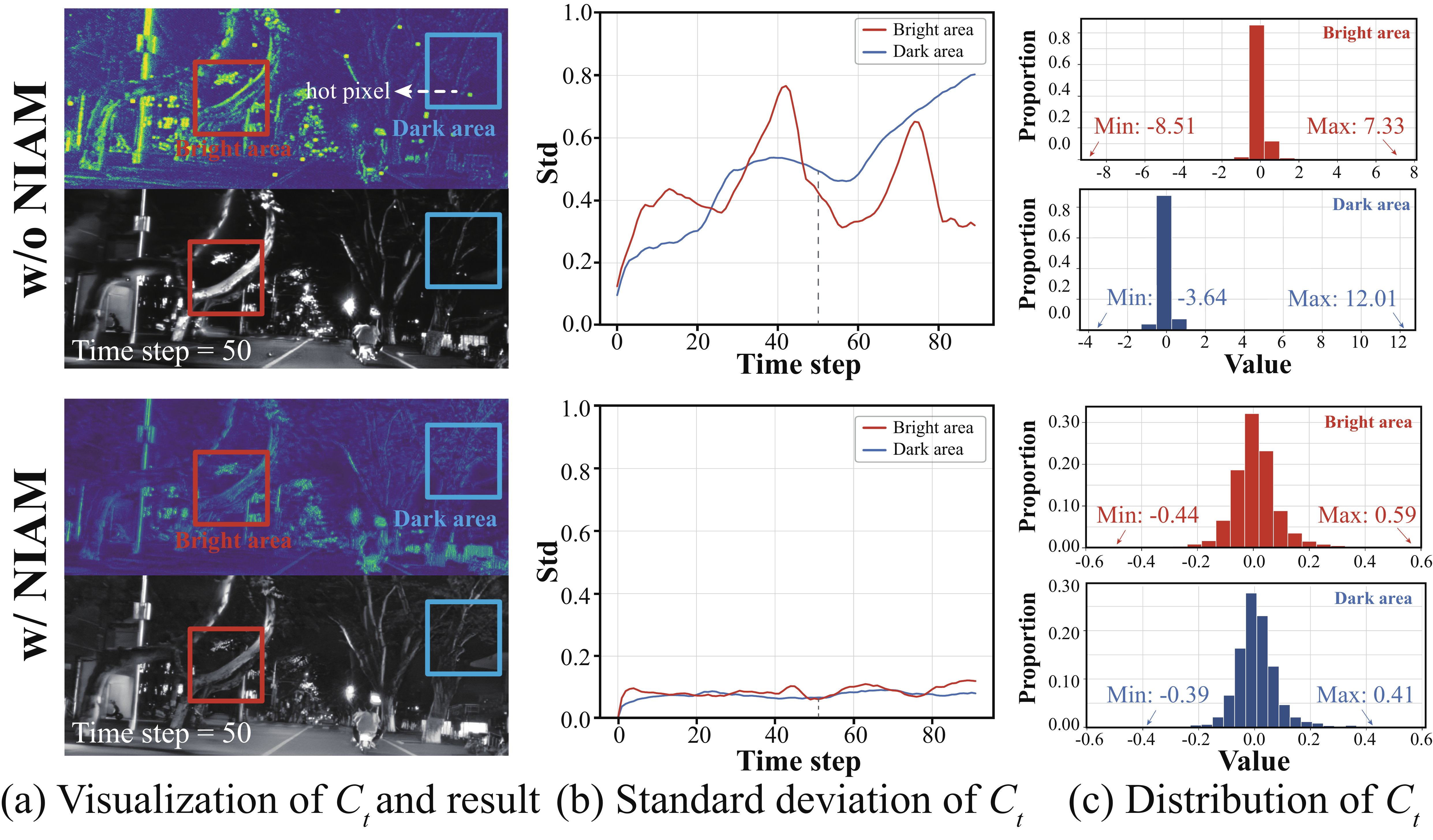}
    \caption{Effectiveness of NIAM. (a) Visualization of memory cell $\mathcal{C}_t^l$ and reconstruction results. (b) Standard deviation of $\mathcal{C}_t^l$ within 90 time steps. (c) The distribution of $\mathcal{C}_t^l$ within 90 time steps.}
    \label{ablation_NIAM}
 \end{figure}

\noindent
\textbf{How does LETC work?}
As shown in Fig.~\ref{ablation_LETC}, the attention of LETC focuses on the correct timestamps as the model iterates. Consequently, the generated voxels exhibit sharper edges. In Table~\ref{tab:ablation_network}, one can see that directly reconstructing ETS data will yield poorer results, as there is a larger gap between the ETS-processed data and real-world data. LETC with trail prior achieves better results through end-to-end training, because the network can learn the mapping from tail-suppressed events to images during the training process.

\noindent
\textbf{Effectiveness of non-uniform illumination awareness.}
We study how NIAM models non-stationary spatiotemporal information flow as shown in Fig.~\ref{ablation_NIAM}. The length of the reconstructed image sequence is 90. For the model w/o NIAM, the standard deviation of $\mathcal{C}_t^l$ exhibits significant fluctuations under non-uniform illumination and noise disturbance (Fig.~\ref{ablation_NIAM} (b)), leading to outliers in temporal propagation as shown in Fig.~\ref{ablation_NIAM} (c), and resulting in contrast distortion in the reconstructed images. NIAM can suppress such abnormal disturbances, and ensure the stability of temporal parameters in both bright and dark areas, thereby reconstructing high-quality images. As shown in Table~\ref{tab:ablation_network}, LAG and SAU play more significant roles in improving the reconstruction quality because they can adaptively perceive changes in event density and stabilize the distribution of events.

\noindent
\textbf{Nighttime dynamic scene imaging: event vs image}.
We capture 6 real-world challenging sequences at nighttime with fast-moving objects and random oscillations of the imaging system. The illumination in the scene ranges between 3 and 20 lux. We carefully set the exposure time of the conventional camera to ensure there was no noticeable motion blur. We compared the results with three SOTA low-light enhancement methods: KinD++ \cite{zhang2021beyond}, SCI \cite{ma2022toward}, and URetinex-Net \cite{wu2022uretinex}. No-reference metrics such as LOE \cite{wang2013naturalness}, NIQE \cite{mittal2012making}, and SPAQ \cite{fang2020perceptual} are adopted for evaluation. Besides, we perform a user study (US) to quantify the subjective visual. We invite 20 human subjects to score the visual quality on a scale from 1 to 10, higher score represents better quality.

\begin{figure}
	\centering
	\subcaptionbox{Events}{\includegraphics[width = 0.33\linewidth]{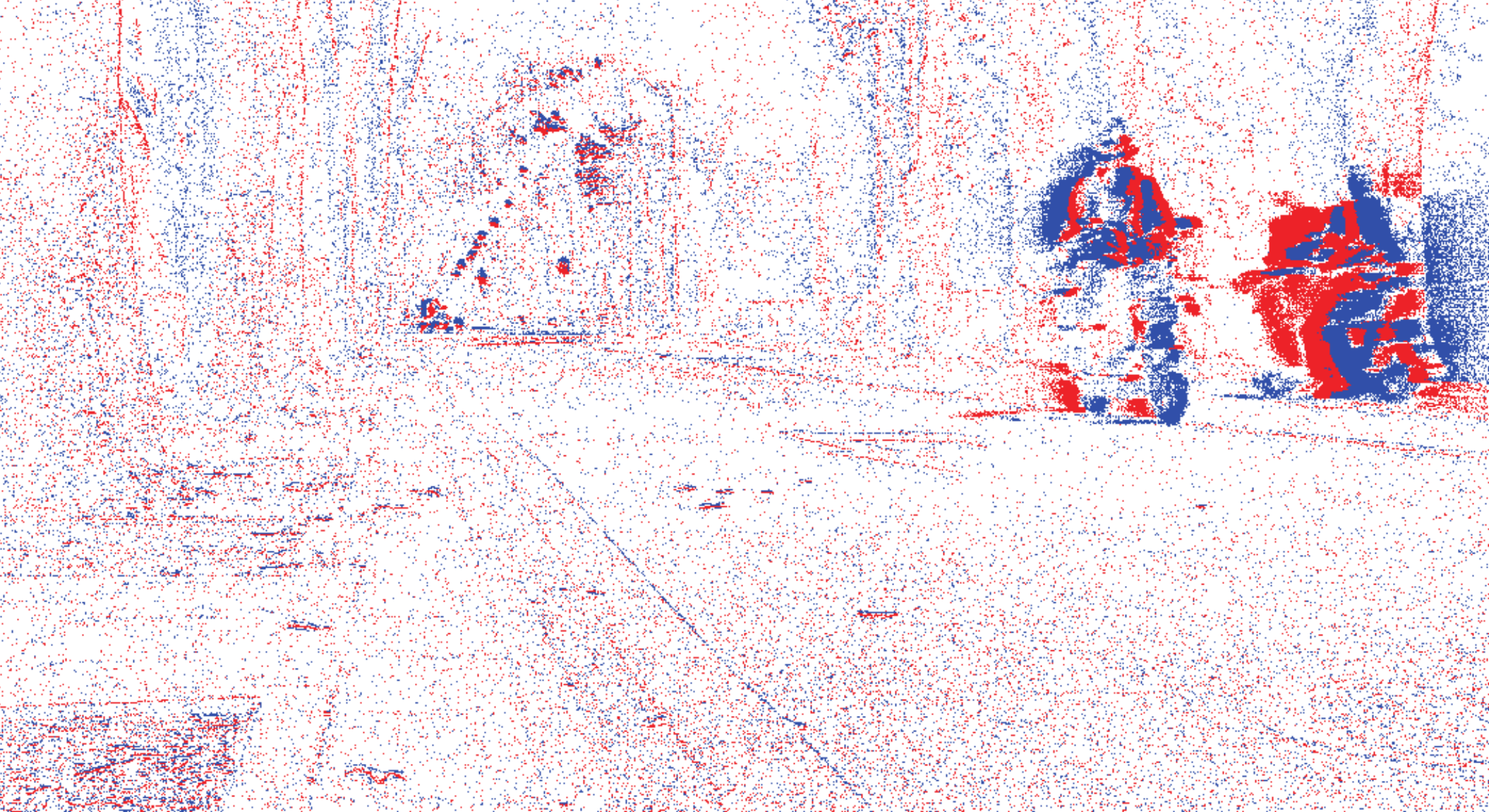}}\hfill
	\subcaptionbox{NER-Net~(Ours)} {\includegraphics[width = 0.33\linewidth]{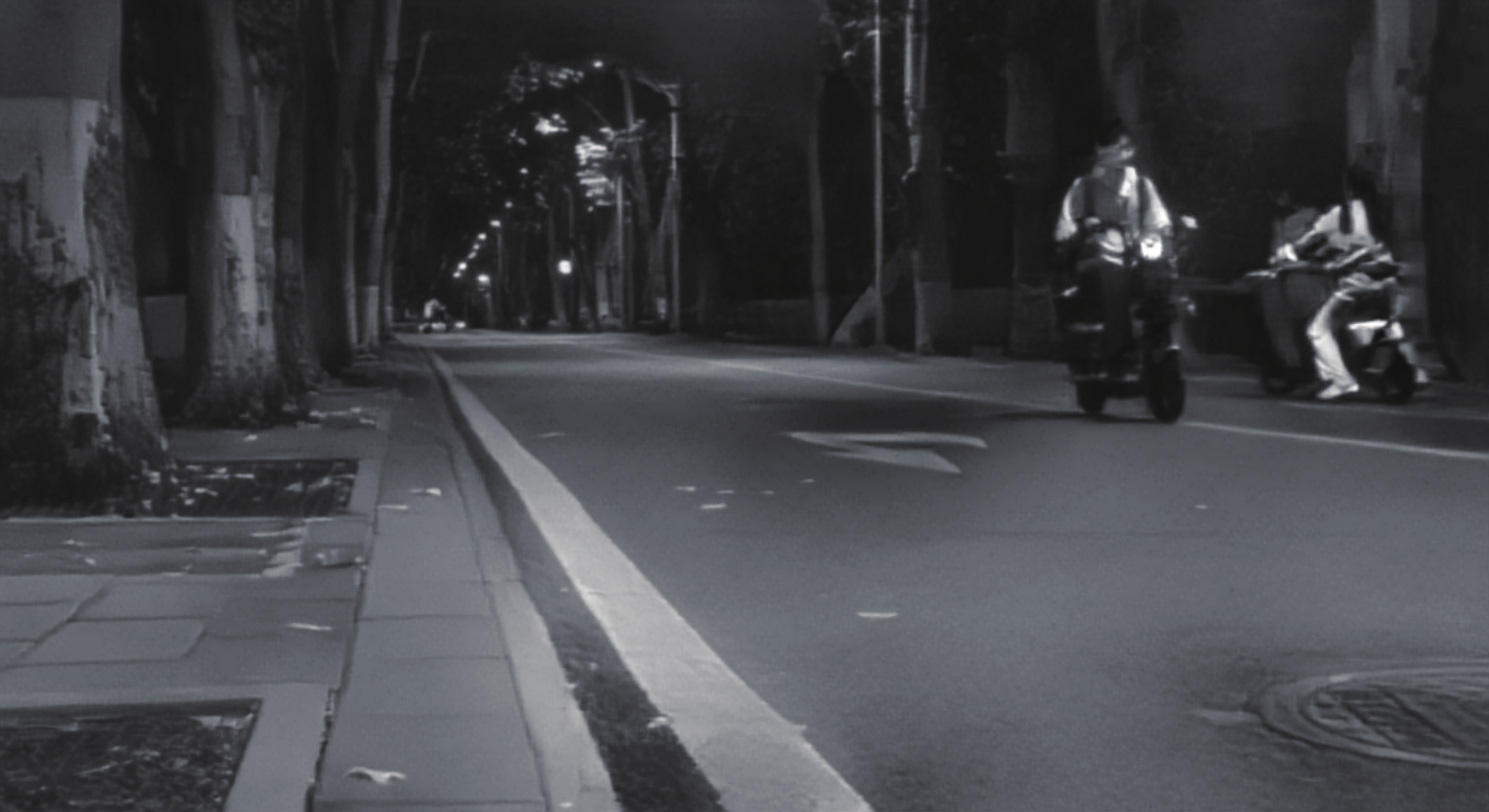}}\hfill
	\subcaptionbox{Image}{\includegraphics[width =0.33\linewidth]{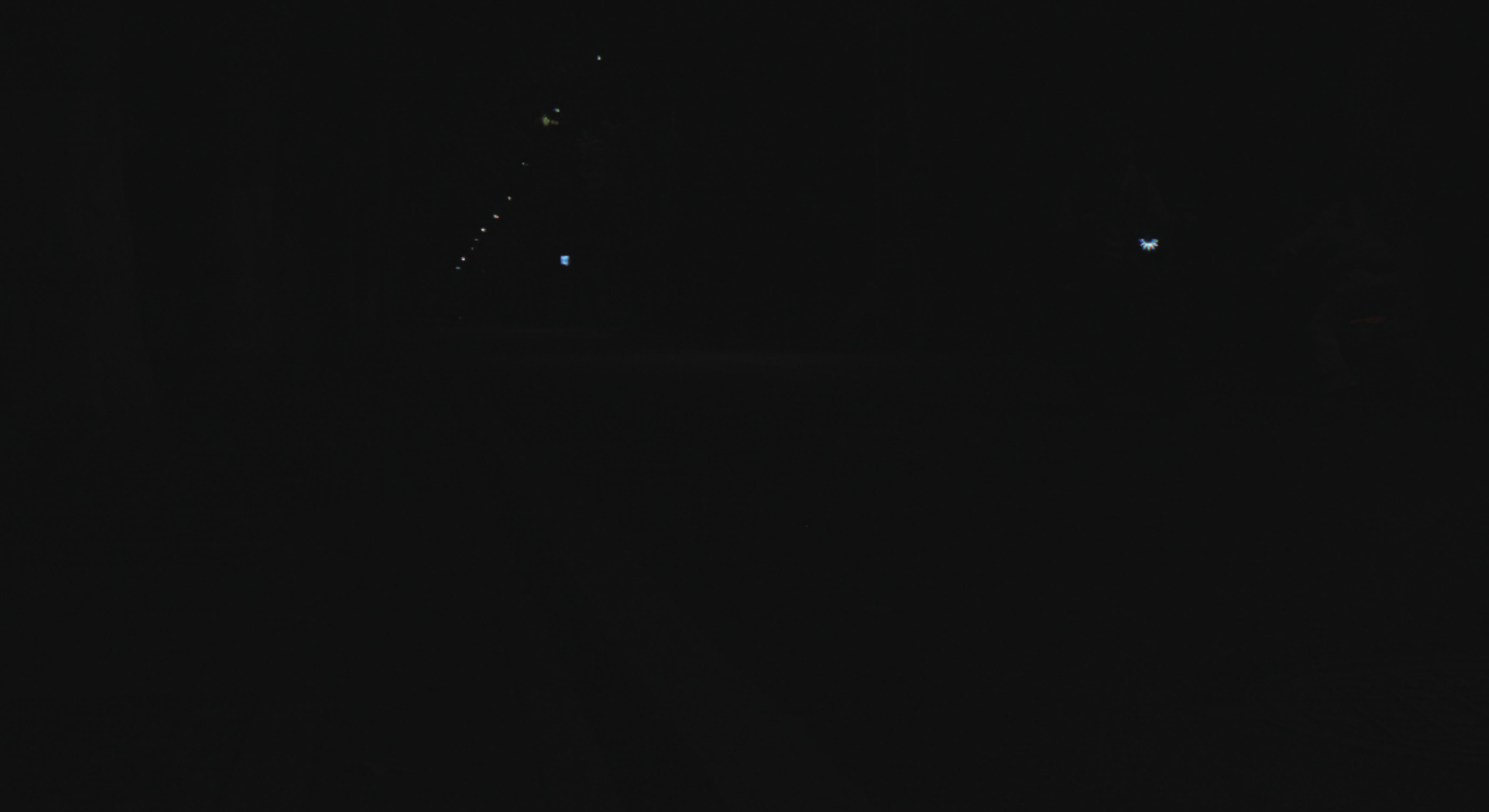}}\\
	\subcaptionbox{KinD++ \cite{zhang2021beyond}}{\includegraphics[width = 0.33\linewidth]{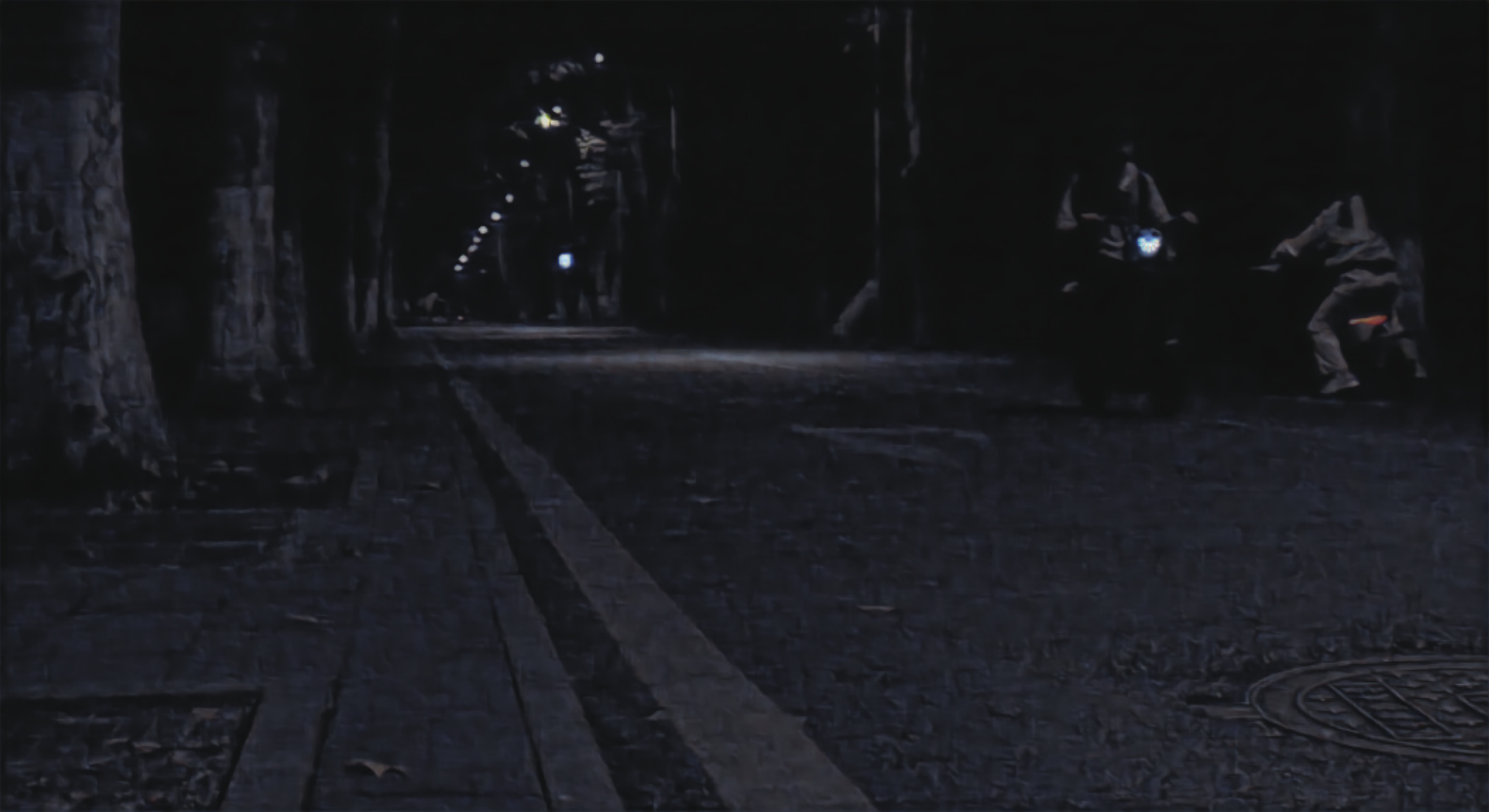}}\hfill
	\subcaptionbox{SCI \cite{ma2022toward}}{\includegraphics[width = 0.33\linewidth]{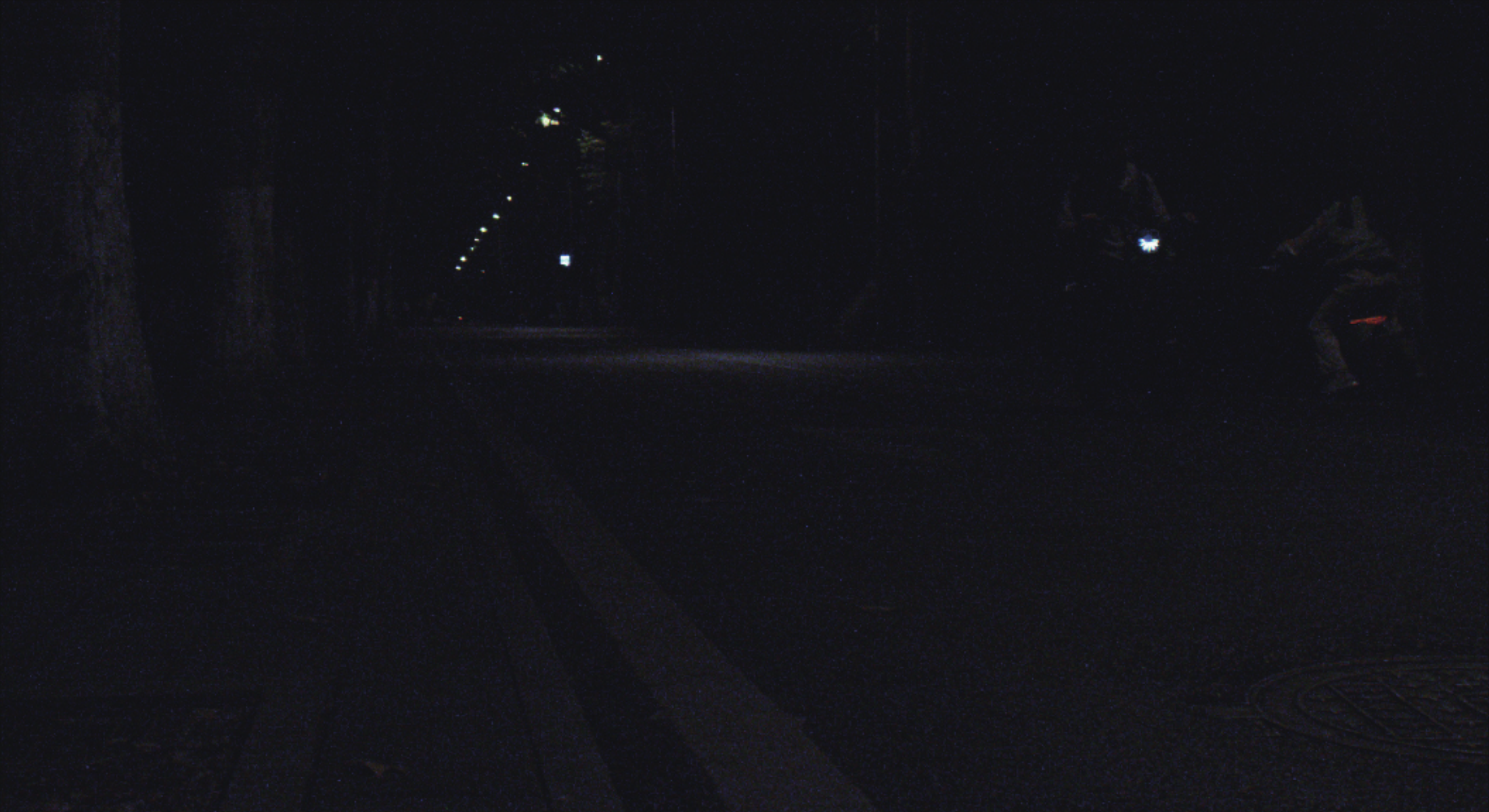}}\hfill
	\subcaptionbox{URetinex-Net \cite{wu2022uretinex}}{\includegraphics[width = 0.33\linewidth]{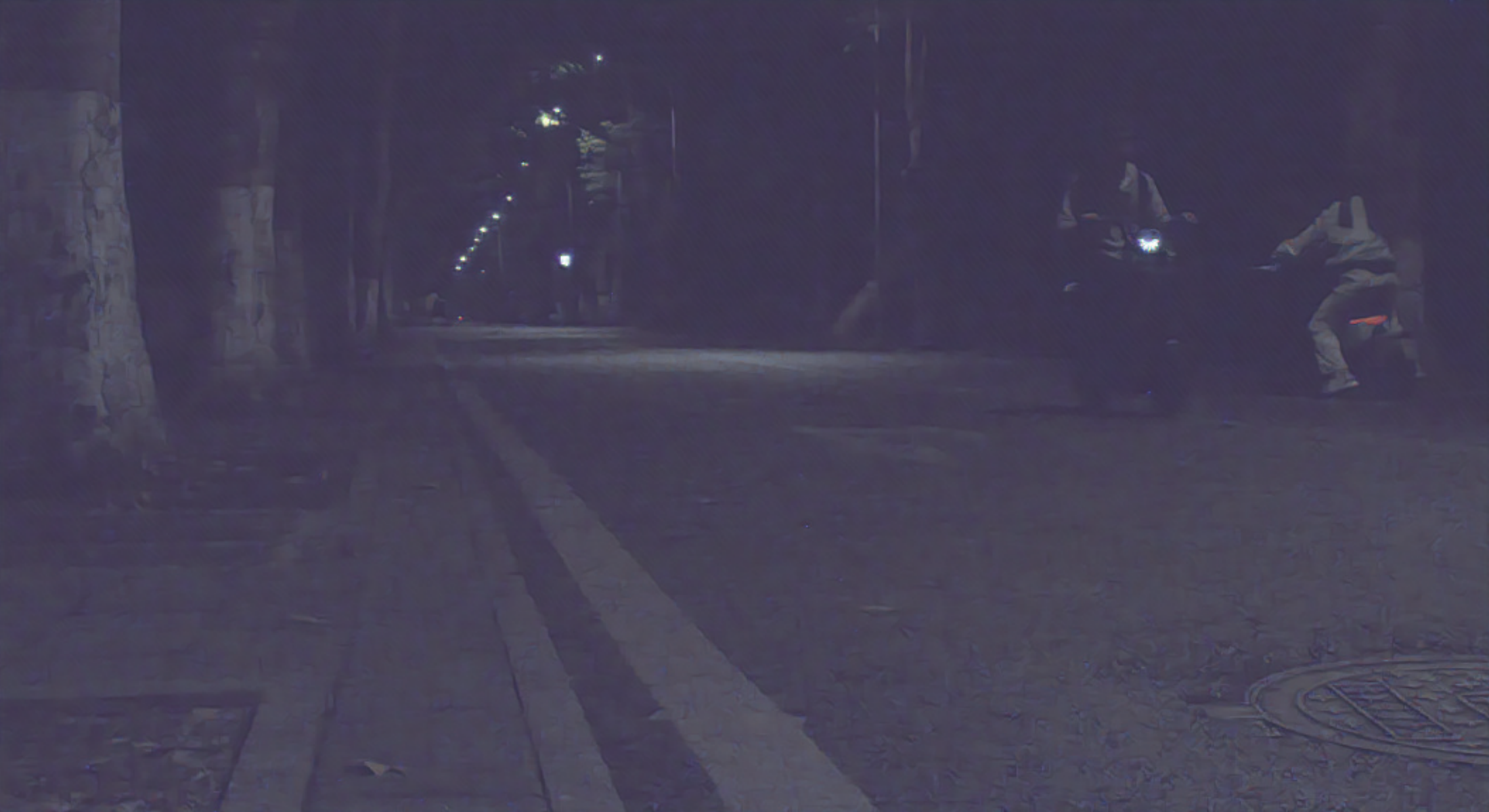}}\\
    \vspace{-0.2cm}
	\caption{Visual comparisons on nighttime dynamic scenes.}
    \vspace{-0.3cm}
  	\label{fig_event_vs_frame}
\end{figure}

Table~\ref{tab:ablation_event_vs_frame} reports that NER-Net achieves the best performance. From Fig.~\ref{fig_event_vs_frame}, it is observed that the excessively short exposure time of conventional cameras used to capture fast-moving objects, leads to the loss of information that is challenging to restore. In contrast, NER-Net can reconstruct scene intensity well with events.

\noindent
\textbf{Limitation.}
NER-Net performs well in the presence of non-uniform artificial lighting and severe sensor noise but fails in extreme low-light scenarios (\eg $<$0.5 lux). The reason is that the drastic reduction of events triggered by object motion under extremely low illumination, it is insufficient for reconstructing reasonable scene details. In addition, the proposed method does not take into account the color information. In the future, we attempt to employ frame-based cameras to provide additional scene information.

\begin{table}
\footnotesize
  \centering
  \setlength{\abovecaptionskip}{5pt}
  \setlength{\belowcaptionskip}{-8pt}

  \begin{tabular}{ccccc}
	\toprule
    Method & LOE~$\downarrow$ & NIQE~$\downarrow$ & SPAQ~$\uparrow$ & US~$\uparrow$  \\
	\midrule
     KinD++ \cite{zhang2021beyond} &113.02  &26.24 & 3708.02 & 5.03 \\
     SCI \cite{ma2022toward} &99.78 &12.49 & 1658.48 & 3.92 \\
     URetinex-Net \cite{wu2022uretinex} &123.45 &13.19 & \textbf{9228.63} & 4.34 \\
     NER-Net(Ours) &\textbf{92.81} &\textbf{11.78} & 8205.78 & \textbf{6.02} \\
	\bottomrule
    \end{tabular}
    \caption{Quantitative results on nighttime dynamic scene.}
 \label{tab:ablation_event_vs_frame}
\end{table}

\section{Conclusion}
In this work, we propose an event-based solution for nighttime dynamic scene imaging and propose the first paired real-world dataset comprising high-quality pixel-aligned GTs for low-light events. Moreover, we demonstrate that the non-stationary status of events under nighttime non-uniform illumination is a key factor degrading the quality of reconstructed images. We model the spatiotemporal disturbance process based on the mechanisms of temporal trailing effects and the spatial non-uniform response of events. The proposed method significantly outperforms the state-of-the-art methods. We believe that our work can contribute to the application of event cameras in real nighttime scenarios.

\noindent
\textbf{Acknowledgments.}
 This work was supported by the National Natural Science Foundation of China under Grant 62371203. The computation is completed in the HPC Platform of Huazhong University of Science and Technology.

\clearpage
\newpage

{
    \small
    \bibliographystyle{ieeenat_fullname}
    \bibliography{main}
}


\end{document}